\documentclass{article}

\PassOptionsToPackage{round,compress}{natbib}
\PassOptionsToPackage{colorlinks,allcolors=blue,backref=page}{hyperref}

\usepackage[frak=boondox]{mathalfa}

\usepackage[preprint]{neurips_2024}

\renewcommand\acksection{\section*{Acknowledgements}\addcontentsline{toc}{section}{Acknowledgements}}

\usepackage[utf8]{inputenc} 
\usepackage[T1]{fontenc}    
\usepackage{hyperref}       
\usepackage{url}            
\usepackage{booktabs}       
\usepackage{amsfonts}       
\usepackage{nicefrac}       
\usepackage{microtype}      
\usepackage{xcolor}         

\usepackage{graphicx}
\usepackage{enumitem}

\usepackage{amsmath}

\usepackage[capitalize,noabbrev]{cleveref}

\renewcommand*{\backref}[1]{}
\renewcommand*{\backrefalt}[4]{%
    \ifcase #1\or Cited on page~#2.\else Cited on pages~#2.\fi%
}

\usepackage{mathtools}
\usepackage{amssymb}

\newcommand\limp{\Rightarrow}
\newcommand\liff{\Leftrightarrow}
\newcommand\Reals{{\mathbb{R}}}
\newcommand\PosReals{{\Reals^{+}}}
\newcommand\Natz{{\mathbb{N}}}
\newcommand\PosNats{{\mathbb{N}^{+}}}

\newcommand\eps{\varepsilon}
\newcommand\set[1]{\left\{#1\right\}}
\newcommand\Set[2]{{\left\{\,#1\,:\,#2\,\right\}}} 
\newcommand\range[2]{{{#1},\ldots,{#2}}}
\newcommand\setrange[2]{{\set{{#1},\ldots,{#2}}}}
\newcommand\subrange[3]{{#3_{#1},\ldots,#3_{#2}}}

\newcommand\sign{\operatorname{sign}}
\newcommand\abs{\operatorname{abs}}
\newcommand\cardinality[1]{{\left|#1\right|}}
\newcommand\argmin{\operatorname{arg\-min}}
\newcommand\argmax{\operatorname{arg\-max}}
\newcommand\modulo{\operatorname{mod}}

\newcommand\W[1][]{\mathcal{W}^{#1}}
\newcommand\U[1][]{\mathcal{U}^{#1}}
\newcommand\prange[4][h]{{\range{#2_1,#3_1,#4_1}{#2_{#1},#3_{#1},#4_{#1}}}}
\newcommand\abcrange[1][h]{\prange[#1]{a}{b}{c}}
\newcommand\rank{\operatorname{rank}}
\newcommand\prank[1]{\operatorname{prank}_{#1}}
\newcommand\Brank[1]{\mathfrak{B}_{#1}}
\newcommand\RS[2][h]{\Xi(#1, #2)}

\newcommand\norm[2]{\left\|{#2}\right\|_{#1}}
\newcommand\dist[3]{\norm{#1}{#2-#3}}
\newcommand\nbhd[3]{\bar{B}_{#1}\!\left(#3;#2\right)}
\newcommand\cnorm[1]{\norm{\infty}{#1}}
\newcommand\cdist[2]{\dist{\infty}{#1}{#2}}
\newcommand\cnbhd[2]{\nbhd\infty{#1}{#2}}

\newcommand\CALL[1]{\textsc{#1}}
\newcommand\BigO{{\mathcal{O}}}
\newcommand\reducto{\rightarrow}
\newcommand\complexityclass[1]{\ensuremath{\mathcal{#1}}}
\newcommand\Poly{\complexityclass{P}}
\newcommand\NP{\complexityclass{NP}}
\newcommand\complexproblem[1]{\texttt{#1}}
\newcommand\SAT{\complexproblem{SAT}}
\newcommand\tSAT{\complexproblem{$3$-SAT}}
\newcommand\ptSAT{\complexproblem{planar $3$-SAT}}
\newcommand\pttSAT{\complexproblem{planar $3$-SAT$_{\bar3}$}}
\newcommand\xSAT{\complexproblem{xSAT}}
\newcommand\PAR{\complexproblem{PR}}
\newcommand\uPAR{\complexproblem{PR-b}}
\newcommand\SSum{\complexproblem{SSum}}
\newcommand\SSZ{\complexproblem{SSZ}}
\newcommand\UPC{\complexproblem{UPC}}
\newcommand\UVCp[1][p]{\ensuremath{\complexproblem{UVC}^{#1}}}
\newcommand\UPP{\complexproblem{UPP}}

\newcommand\usgCP{\complexproblem{usgCP}}

\usepackage{amsthm} 
\usepackage{thmtools}
\usepackage{thm-restate}

\declaretheorem[parent=section]{theorem}
\declaretheorem[sibling=theorem]{lemma}
\declaretheorem[sibling=theorem]{proposition}
\declaretheorem[sibling=theorem]{corollary}

\declaretheorem[sibling=theorem,style=definition]{remark}
\declaretheorem[sibling=theorem,style=definition]{example}
\declaretheoremstyle[
    headfont=\normalfont\itshape,
    qed=\qedsymbol,
]{proofstyle}
\declaretheorem[
    name={Proof},
    style=proofstyle,
    unnumbered
]{myproof}
\declaretheoremstyle[
    headfont=\normalfont\itshape,
    qed={$\lozenge$},
]{proofsketchstyle}
\declaretheorem[
    name={Proof sketch},
    style=proofsketchstyle,
    unnumbered,
]{proofsketch}
\declaretheorem[parent=section,style=definition]{algorithm}
\usepackage[noEnd=false,indLines=false,commentColor=gray]{algpseudocodex}
\declaretheoremstyle[
    notebraces={}{},
    notefont=\bfseries,
    headformat=\NAME \NOTE,
]{problemstyle}
\declaretheorem[style=problemstyle,name=Problem]{problem}
\newcommand{\pref}[2]{Problem \hyperref[#1]{#2}}
\newcommand{\Pref}[2]{Problem \hyperref[#1]{#2}}
\newcommand\probUPC{\pref{prob:upc}{\UPC}}
\newcommand\probUPP{\pref{prob:upp}{\UPP}}
\newcommand\probxSAT{\pref{prob:xsat}{\xSAT}}
\newcommand\probPAR{\pref{prob:par}{\PAR}}
\newcommand\probUPAR{\pref{prob:upar}{\uPAR}}
\newcommand\probSS{\pref{prob:ssum}{\SSum}}
\newcommand\probSSZ{\pref{prob:ssumzero}{\SSZ}}

\title{%
    Proximity to Losslessly Compressible Parameters
}

\author{%
  Matthew Farrugia-Roberts \\
  School of Computing and Information Systems \\
  The University of Melbourne \\
  \texttt{matthew@far.in.net}
}

\begin{document}

\maketitle

\begin{abstract}
    To better understand complexity in neural networks, we theoretically
    investigate the idealised phenomenon of lossless network compressibility,
    whereby an identical function can be implemented with fewer hidden units.
    In the setting of single-hidden-layer hyperbolic tangent networks,
    we define the rank of a parameter as the minimum number of hidden units
    required to implement the same function.
    We give efficient formal algorithms for optimal lossless compression
    and computing the rank of a parameter.
    Losslessly compressible parameters are atypical, but their existence has
    implications for nearby parameters.
    We define the proximate rank of a parameter as the rank of the most
    compressible parameter within a small $L^\infty$ neighbourhood.
    We give an efficient greedy algorithm for bounding the proximate rank of
    a parameter, and show that the problem of tightly bounding the proximate
    rank is \NP-complete.
    These results lay a foundation for future theoretical and empirical work
    on losslessly compressible parameters and their neighbours.
\end{abstract}


\section{Introduction}
\label{sec:intro}

Learned neural networks are often simpler than parameter counting would
suggest.
Architectures used in practice can easily memorise randomly labelled data
\citep{Zhang+2017,Zhang+2021}.
Yet, they tend to learn simple functions that are \emph{approximately
compressible}, in that there are smaller networks implementing similar
functions
    \citep[e.g.,][]{Bucilua+2006,Hinton+2015,DistilBERT2020}.

To advance our understanding of neural network complexity, we propose
studying the idealised phenomenon of \emph{lossless compressibility} of
neural network parameters, whereby an \emph{identical} function can be
implemented by some network with fewer units.
Such parameters are idealised examples of parsimonious solutions to learning
problems.
Moreover, their neighbouring parameters implement similar functions, and are
therefore examples of approximately compressible networks.

In this paper, we study losslessly compressible parameters and their
neighbourhoods in the setting of single-hidden-layer hyperbolic tangent
networks.
While this setting is not immediately relevant to modern deep learning,
much of our analysis applies to any nonlinearity and to any individual
feed-forward layer of a larger architecture
    (see \cref{sec:discussion}).
We therefore offer the following theoretical contributions as
a first step towards understanding lossless compressibility in modern
architectures.
\begin{enumerate}
    \item
        In \cref{sec:rank}, we give efficient formal algorithms for
        optimal lossless compression of single-hidden-layer hyperbolic
        tangent networks,
        and for computing the \emph{rank} of a parameter---the minimum
        number of hidden units required to implement the same function.
    
    \item
        In \cref{sec:prank}, we turn to studying the neighbours of losslessly
        compressible parameters.
        We define the \emph{proximate rank} of a parameter as the rank of the
        most compressible parameter within a small $L^\infty$ neighbourhood.
        We give a greedy algorithm for bounding this value.

    \item
        In \cref{sec:complexity}, we show that
            detecting proximity to parameters with a given maximum rank
            (equivalently, bounding the proximate rank below a given value)
        is an \NP-complete decision problem.
        It follows that computing the proximate rank exactly is \NP-hard.
\end{enumerate}
While the latter can be seen as a somewhat negative result, the proof also
establishes connections between the problem of computing proximate rank and
well-studied hard decision problems, opening the door to a rich literature on
tractable approximations in practical instances.

Our contributions lay a foundation for future theoretical and empirical work
characterising lossless compressibility in modern architectures and detecting
proximity to losslessly compressible parameters in learned neural networks.
This research direction could offer a new lens on the phenomenon of
approximate compressibility in deep learning.
We discuss this research direction in \cref{sec:discussion}.

\section{Related work}
\label{sec:related}

We summarise work related to the topic of lossless compressibility of neural
networks. To the best of our knowledge, no prior work studies proximity to
losslessly compressible parameters. We defer discussion of related work in
computational complexity theory to \cref{sec:complexity,apx:upc}.

\paragraph{Approximate compression.}

There is a sizeable empirical literature on approximate compression in neural
networks, including via pruning, quantisation, and distillation
    \citetext{%
        see \citealp{Cheng+2018,Cheng+2020} or \citealp{Choudhary+2020} for
        an overview%
    }.
Approximate compressibility has also been proposed as a learning objective
    \citep{Hinton+vanCamp1993,Aytekin+2019}
and used in deriving generalisation bounds
    \citep{Suzuki+2020a,Suzuki+2020b}.
We adopt a convention of measuring network size by counting units, noting
that this is one of various alternative conventions in the literature.

\paragraph{Lossless compression.}

There has been less work on \emph{lossless} compression, requiring the
network's outputs to be exactly preserved.
\citet{Serra+2020} give a partial lossless compression algorithm for
multi-layer ReLU networks that preserves the implemented function over some
input domain.
Their algorithm exploits \emph{some} opportunities for removing units, but
does not claim to (and does not in general) \emph{optimally} compress
networks.
By contrast, we give an algorithm for optimal lossless compression over all
inputs in the simpler setting of single-hidden-layer hyperbolic tangent
networks.

\paragraph{Functional equivalence.}

For single-hidden-layer hyperbolic tangent networks, \citet{Sussmann1992}
showed that, for almost all parameters, two parameters implement identical
functions if and only if they are related by simple operations of exchanging
or negating the weights of hidden units.
Similar results have been shown for various architectures, including
architectures with different nonlinearities
    \citep{Albertini+1993,Kurkova+Kainen1994,Phuong+Lampert2020},
multiple hidden layers
    \citep{Chen+1993,Fefferman+Markel1993,Fefferman1994,Phuong+Lampert2020},
and more complex connection graphs
    \citep{Vlacic+Bolcskei2021,Vlacic+Bolcskei2022}.

Lossless compressibility is precisely the existence of functionally equivalent
parameters with fewer units.
The function-preserving operations cited above generally preserve the
number of units.\footnote{%
    It follows that losslessly compressible parameters occupy a measure
    zero subset of parameter space.
    We note that learning exerts a non-random selection pressure, so these
    parameters may still be relevant in practice.
}
\citet{Farrugia2023} studies additional function-preserving operations
available for losslessly compressible single-hidden-layer hyperbolic tangent
parameters, giving an algorithm for identifying pairs of functionally
equivalent parameters that achieves optimal lossless compression as a
side-effect.
We give a more efficient optimal lossless compression algorithm.

\paragraph{Lossless expansion.}

There is also work adopting a dual perspective of cataloguing various ways of
\emph{adding} hidden units to a neural network while exactly preserving the
implemented function.
\citet{Fukumizu+Amari2000} and \citet{Fukumizu+2019} show that some of the
resulting losslessly compressible parameters are critical points of the loss
landscape.
\citet{Simsek+2021} and \citet{Farrugia2023} show that sets of equivalent
losslessly compressible parameters have a rich structure that reaches
throughout the parameter space.
We show a similar result for the set of all sufficiently losslessly
compressible (not necessarily equivalent) parameters in \cref{apx:brank}.

\paragraph{Information singularities.}

Losslessly compressible parameters are singularities in the Fisher
information landscape
    \citep{Fukumizu1996},
and if they are critical points of the loss landscape they are degenerate.
This makes them highly relevant to singular statistical theories of deep
learning
    \citep{greybook,goodpaper}.
For example, these singularities influence learning dynamics in their
neighbourhood
    \citep{Amari+2006,Wei+2008,Cousseau+2008,Amari+2018}.

\section{Preliminaries}
\label{sec:prelims}

\paragraph{Architecture.}

We consider a family of fully-connected, feed-forward neural network
architectures with one input unit, one biased output unit, and one hidden
layer of $h \in \Natz$ biased hidden units with the hyperbolic tangent
nonlinearity
        $\tanh(z) = (e^{z} - e^{-z}) / (e^{z} + e^{-z})$.
The weights and biases of the network are encoded in a parameter vector in
the format
    $w = (\abcrange, d) \in \W_h = \Reals^{3h+1}$,
where
    for each hidden unit $i = \range1h$ there is
        an \emph{outgoing weight} $a_i \in \Reals$,
        an \emph{incoming weight} $b_i \in \Reals$,
        and a \emph{bias} $c_i \in \Reals$;
    and $d \in \Reals$ is
        the \emph{output unit bias}.
Thus each parameter $w \in \W_h$ indexes a mathematical function
    $f_w : \Reals \to \Reals$
such that
    $f_w(x) = d + \sum_{i=1}^h a_i \tanh(b_i x + c_i)$.
All of our results generalise to networks with multi-dimensional inputs and
outputs (see \cref{apx:multidim}).

\paragraph{Reducibility.}

Two parameters $w \in \W_h$ and $w' \in \W_{h'}$ are \emph{functionally
equivalent} if $f_w = f_{w'}$ as functions on $\Reals$
    ($\forall x \in \Reals, f_w(x) = f_{w'}(x)$).
A parameter $w \in \W_h$ is \emph{(losslessly) compressible} if and only if
$w$ is functionally equivalent to some
    $w' \in \W_{h'}$ with fewer hidden units $h' < h$
(otherwise, $w$ is \emph{incompressible}).
\citet{Sussmann1992} showed that a simple condition,
    \emph{reducibility,}
is necessary and sufficient for lossless compressibility.
A parameter $(\abcrange, d) \in \W_h$ is \emph{reducible} if and only if
it satisfies any of the following \emph{reducibility conditions}:
\begin{enumerate}[label=(\roman*)]
    \item\label{item:reducibility:1}
        $a_i = 0$ for some $i$, or
    \item\label{item:reducibility:2}
        $b_i = 0$ for some $i$, or
    \item\label{item:reducibility:3}
        $(b_i, c_i) = (b_j, c_j)$ for some $i \neq j$, or
    \item\label{item:reducibility:4}
        $(b_i, c_i) = (-b_j, -c_j)$ for some $i \neq j$.
\end{enumerate}

Each reducibility condition suggests a simple operation to remove a hidden
unit while preserving the function \citep{Sussmann1992, Farrugia2023}.
\begin{enumerate}[label=(\roman*)]
    \item
        Units with zero outgoing weight contribute zero to the function, and
        can be \emph{eliminated}, that is, can be removed from the network.
    \item
        Units with zero incoming weight contribute a constant to the
        function, and can also be \emph{eliminated} after incorporating the
        constant into the output bias.
    \item
        Unit pairs with identical incoming weight and bias contribute
        proportionally to the function, and can be \emph{merged} into a
        single unit with the sum of their outgoing weights as the new
        outgoing weight
        (for a net removal of one unit).
    \item
        Unit pairs with identically negative incoming weight and bias also
        contribute in proportion, since the hyperbolic tangent is odd.
        Such pairs can also be \emph{merged} into a single unit with the
        difference of their outgoing weights as the new outgoing weight.
\end{enumerate}

\paragraph{Uniform neighbourhood.}

The \emph{uniform norm} (or \emph{$L^\infty$ norm}) of a vector
    $v\in\Reals^p$
is the largest absolute component of $v$:
    $\cnorm{v} = \max_{i=1}^p \abs(v_i)$.
The \emph{uniform distance} between $v$ and $u \in \Reals^p$ is
    $\cdist{u}{v}$.
Given a positive \emph{radius} $\eps\in\PosReals$, the
    \emph{(closed) uniform neighbourhood of $v$}
is the set of vectors of uniform distance at most $\eps$ from $v$:
    $\cnbhd\eps{v} = \Set{u\in\Reals^p}{\cdist{u}{v}\leq\eps}$.

\paragraph{Computational complexity theory.}

We informally review several basic notions from computational
complexity theory.\footnote{%
    We refer readers to \citet{Garey+Johnson1979} for a rigorous introduction
    to the requisite computational complexity theory for our results
    (in terms of formal languages, encodings, and Turing machines).
}
A \emph{decision problem} is a tuple $(I, J)$ where $I$ is a set of
\emph{instances} and $J \subseteq I$ is a subset of \emph{affirmative
instances}.
A \emph{solution} is a deterministic algorithm that determines if any given
instance $i \in I$ is affirmative ($i\in J$).
\Poly{} is the class of decision problems with polytime solutions (run-time
polynomial in the instance size).
\NP{} is the class of decision problems for which a deterministic polytime
algorithm can verify affirmative instances given a certificate.

A \emph{reduction} from one decision problem $X = (I, J)$ to another
$Y = (I', J')$ is a deterministic polytime algorithm implementing a mapping
$\varphi : I \to I'$ such that $\varphi(i) \in J' \liff i \in J$.
If such a reduction exists, we say \emph{$X$ is reducible to $Y$} and write
$X \reducto Y$.\footnote{%
    Context should suffice to distinguish reducibility \emph{between decision
    problems} and \emph{of network parameters}.
}
Reducibility is transitive.
A decision problem $Y$ is \emph{\NP-hard} if all problems in $\NP$ are
reducible to $Y$ ($\forall X \in \NP, X \reducto Y$).
$Y$ is \emph{\NP-complete} if $Y$ is \NP-hard and $Y \in \NP$.

\section{Optimal lossless compression and rank}
\label{sec:rank}

We consider the problem of lossless neural network compression: finding,
given a compressible parameter, a functionally equivalent but incompressible
parameter.
The following algorithm solves this problem by eliminating units meeting
reducibility conditions~\ref{item:reducibility:1} or~\ref{item:reducibility:2}
and merging unit pairs meeting reducibility
conditions~\ref{item:reducibility:3} or~\ref{item:reducibility:4} in ways
preserving functional equivalence.

\begin{algorithm}[Lossless neural network compression]
    \label{algo:lossless-compression}
    Given $h\in\Natz$, proceed:
    \begin{algorithmic}[1]
    \Procedure{Compress}{$w=(\abcrange,d)\in\W_h$}
        \LComment{Stage 1: Eliminate units with incoming weight zero
            (incorporate into new output bias $\delta$)}
        \State $I \gets \Set{i\in\setrange1h}{b_i \neq 0}$
        \State $\delta \gets d + \sum_{i \notin I} \tanh(c_i)\cdot a_i$
        \LComment{Stage 2: Partition and merge remaining units by incoming
            weight and bias}
        \State $\subrange1J\Pi \gets$
            partition $I$ by the value of $\sign(b_i) \cdot (b_i,c_i)$
        \For{$j \gets \range1J$}
            \State $\alpha_j \gets \sum_{i\in\Pi_j} \sign(b_i) \cdot a_i$
            \State $\beta_j, \gamma_j \gets
                \sign(b_{\min \Pi_j}) \cdot (b_{\min \Pi_j}, c_{\min \Pi_j})$
        \EndFor
        \LComment{Stage 3: Eliminate merged units with outgoing weight zero}
        \State $k_1, \ldots, k_r \gets \Set{j\in\setrange1J}{\alpha_j \neq 0}$
        \LComment{Construct a new parameter with the remaining merged units}
        \State \Return $(
                        \alpha_{k_1}, \beta_{k_1}, \gamma_{k_1},
                        \ldots,
                        \alpha_{k_r}, \beta_{k_r}, \gamma_{k_r},
                        \delta
                    ) \in \W_r$
    \EndProcedure
    \end{algorithmic}
\end{algorithm}

\begin{restatable}[\cref{algo:lossless-compression} correctness]{theorem}%
    {thmAlgoLosslessCompression}\label{thm:algo:lossless-compression}
    Given $w \in \W_h$, compute $w' = \CALL{Compress}(w) \in \W_r$.
    (i)~$f_{w'} = f_w$, and (ii)~$w'$ is incompressible.
\end{restatable}

\begin{proofsketch}[Full proof in \cref{apx:proofs}]
    For (i), note that
        units eliminated in Stage~1 contribute a constant $a_i\tanh(c_i)$,
        units merged in Stage~2 have proportional contributions
            ($\tanh$ is odd),
        and merged units eliminated in Stage~3 do not contribute.
    For (ii), by construction, $w'$ satisfies no reducibility conditions,
    so $w'$ is not reducible and is thus incompressible by \citet{Sussmann1992}.
\end{proofsketch}

We define the \emph{rank}\footnotemark{} of a neural network parameter $w \in
\W_h$,
    denoted $\rank(w)$,
as the minimum number of hidden units required to implement $f_w$:
    $
      \rank(w) = \min\Set{h' \in \Natz}{\exists w' \in \W_{h'};\ f_w = f_{w'}}
    $.
The rank is also the number of hidden units in $\CALL{Compress}(w)$, since
\cref{algo:lossless-compression} produces an incompressible parameter.
Computing the rank is therefore a trivial matter of counting the units
after performing lossless compression.
The following is a streamlined algorithm, following
    \cref{algo:lossless-compression}
but removing steps that don't influence the final count.

\footnotetext{%
    In the multi-dimensional case
        (see \cref{apx:multidim}),
    our notion of rank generalises the familiar notion from linear algebra,
    where the rank of a linear transformation corresponds to the minimum
    number of hidden units required to implement the transformation with an
    unbiased linear neural network
        \citep[cf.][]{Piziak+Odell1999}.
    Unlike in the linear case, our non-linear rank is not bounded by
    the input or output dimensionality.
}

\begin{algorithm}[Rank of a neural network parameter]\label{algo:rank}
    Given $h \in \Natz$, proceed:
    \begin{algorithmic}[1]
    \Procedure{Rank}{$w=(\abcrange,d)\in\W_h$}
        \LComment{
            Stage 1: Identify units with incoming weight nonzero (not
            eliminated)
        }
        \State $I \gets \Set{i\in\setrange1h}{b_i \neq 0}$
        \LComment{
            Stage 2: Partition these units and compute outgoing weights
            for would-be merged units
        }
        \State $\Pi_1, \ldots, \Pi_J \gets$
                partition $I$ by the value of 
                $\sign(b_i)\cdot(b_i,c_i)$
        \State $\alpha_j \gets \sum_{i\in\Pi_j} \sign(b_i) \cdot a_i$
            \textbf{for} $j \gets \range1J$
        \LComment{Stage 3: Count merged units with outgoing weight nonzero}
        \State \Return $\cardinality{\Set{j\in\setrange1J}{\alpha_j\neq0}}$
            \Comment{$\cardinality{S}$ denotes set cardinality}
    \EndProcedure
    \end{algorithmic}
\end{algorithm}

\begin{theorem}[\cref{algo:rank} correctness]
    Given $w \in \W_h$, $\rank(w) = \CALL{Rank}(w)$.
\end{theorem}

\begin{myproof}
    Let $r$ be the number of hidden units in $\CALL{Compress}(w)$.
    Then $r = \rank(w)$ by \cref{thm:algo:lossless-compression}.
    Moreover, comparing
        \cref{algo:rank,algo:lossless-compression},
    observe $\CALL{Rank}(w) = r$.
\end{myproof}

\begin{remark}
    Both \cref{algo:lossless-compression,algo:rank} require
        $\BigO(h \log h)$ time
    if the partitioning step is performed by first sorting the units by
    lexicographically non-decreasing $\sign(b_i) \cdot (b_i, c_i)$.
\end{remark}

The reducibility conditions characterise the set of parameters $w \in \W_h$
with $\rank(w) \leq h-1$. In \cref{apx:brank} we characterise the set of
parameters with an arbitrary rank bound.

\clearpage

\section{Proximity to low-rank parameters}
\label{sec:prank}

Given a neural network parameter $w \in \W_h$ and a positive radius
$\eps \in \PosReals$, we define the \emph{proximate rank} of $w$ at radius
$\eps$, denoted $\prank{\eps}(w)$, as the rank of the lowest-rank parameter
within a closed uniform ($L^\infty$) neighbourhood of $w$ with radius $\eps$.
That is,
\begin{equation*}
    \prank{\eps}(w) = \min \Set{\rank(u)\in\Natz}{u \in \cnbhd\eps{w}}.
\end{equation*}
The proximate rank measures the proximity of $w$ to the set of parameters
with a given rank bound, that is, sufficiently losslessly compressible
parameters.
We characterise the set of bounded-rank parameters in \cref{apx:brank}, and
we explore some basic properties of the proximate rank in \cref{apx:prank}.

The following greedy algorithm computes an upper bound on the proximate rank.
The algorithm replaces each of the three stages of \cref{algo:rank}
with a relaxed version, as follows.
\begin{enumerate}
    \item
        Instead of eliminating units with zero incoming weight,
        eliminate units with \emph{near} zero incoming weight
            (there is a nearby parameter where these are zero).
    \item
        Instead of partitioning the remaining units by
            $\sign(b_i) \cdot (b_i, c_i)$, 
        \emph{cluster} them by \emph{nearby}
            $\sign(b_i) \cdot (b_i, c_i)$
        (there is a nearby parameter where they have the same 
            $\sign(b_i) \cdot (b_i, c_i)$).
    \item
        Instead of eliminating merged units with zero outgoing weight,
        eliminate merged units with \emph{near} zero outgoing weight
            (there is a nearby parameter where these are zero).
\end{enumerate}
Step (2) is non-trivial, we use a greedy approach, described separately as
    \cref{algo:greedy-approx-partition}.

\begin{algorithm}[Greedy bound for proximate rank]
    \label{algo:greedy-prank-bound}
    Given $h \in \Natz$, proceed:
    \begin{algorithmic}[1]
    \Procedure{Bound}{%
        $\eps\in\PosReals$,
        $w=(\abcrange,d)\in\W_h$%
    }
        \LComment{Stage 1: Identify units with incoming weight not near zero}
        \State $I \gets \Set{i\in\setrange1h}{\cnorm{b_i} > \eps}$
        \LComment{Stage 2: Compute outgoing weights for nearly-mergeable units}
        \State $\subrange1J\Pi
            \gets
            \Call{ApproxPartition}{%
                \eps,\ 
                \sign(b_i)\cdot(b_i,c_i)
                \textbf{ for }i\in I}
            $
            \Comment{\cref{algo:greedy-approx-partition}}
        \State $\alpha_j \gets \sum_{i\in\Pi_j} \sign(b_i) \cdot a_i$
            \textbf{for} $j \gets \range1J$
        \LComment{Stage 3: Count nearly-mergeable units with outgoing weight
            not near zero}
        \State \Return $\cardinality{
                \Set
                    {j\in\setrange1J}
                    {\cnorm{\alpha_j} > \eps\cdot\cardinality{\Pi_j}}
            }$
            \Comment{$\cardinality{S}$ denotes set cardinality}
    \EndProcedure
    \end{algorithmic}
\end{algorithm}

\begin{algorithm}[Greedy approximate partition]
    \label{algo:greedy-approx-partition}
    Given $h \in \Natz$, proceed:
    \begin{algorithmic}[1]
    \Procedure{ApproxPartition}{$\eps\in\PosReals$, $\subrange1hu\in\Reals^2$}
        \State $J \gets 0$
        \For{$i \gets \range1h$}
            \If{\textbf{for some} $j\in\setrange1J$, $\cdist{u_i}{v_j}\leq\eps$}
                \State $\Pi_j \gets \Pi_j \cup \set{i}$
                            \Comment{If near a group-starter, join that group.}
            \Else
                \State $J, v_{J+1}, \Pi_{J+1} \gets J + 1, u_i, \set{i}$
                            \Comment{Else, start a new group with this vector.}
            \EndIf
        \EndFor
        \State \Return $\Pi_1, \ldots, \Pi_J$
    \EndProcedure
    \end{algorithmic}
\end{algorithm}

\begin{restatable}[\cref{algo:greedy-prank-bound} correctness]{theorem}%
    {thmAlgoGreedyPrankBound}\label{thm:algo:greedy-prank-bound}
    For $w\in\W_h$ and $\eps\in\PosReals$,
    $\prank\eps(w) \leq \CALL{Bound}(\eps, w)$.
\end{restatable}

\begin{proofsketch}[Full proof in \cref{apx:proofs}]
    Trace the algorithm to construct a parameter $u \in \W_h$ from the input
    parameter $w$ as follows.
    During Stage~1, set the nearly-eliminable incoming weights to zero.
    Use the group-starting vectors $\subrange1Jv$ from
        \cref{algo:greedy-approx-partition}
    to construct mergeable incoming weights and biases during Stage~2.
    During Stage~3, subtract or add a fraction of the merged unit outgoing
    weight from the outgoing weights of the original units.
    
    Running \cref{algo:rank} on $u$ results in the same output
        $\CALL{Rank}(u) = \CALL{Bound}(\eps,w)$,
    because in the construction of $u$ we have inverted the relaxation of
    each step of \cref{algo:rank} that transformed \cref{algo:rank} into
    \cref{algo:greedy-prank-bound}.
    Therefore, $\rank(u) = \CALL{Bound}(\eps,w)$.

    Moreover, each of the steps of the construction of $u$ from $w$ changes a
    component of $w$ by at most $\eps$, so the resulting parameter $u$ is
    within the uniform neighbourhood $\cnbhd\eps{w}$.
    Therefore, $\prank\eps(w) \leq \rank(u)$.
\end{proofsketch}

\begin{remark}
    \label{remark:algo:greedy-approx-partition}
    Both \cref{algo:greedy-approx-partition,algo:greedy-prank-bound}
    have worst-case runtime complexity $\BigO(h^2)$.
\end{remark}

\newpage

\section{Computational complexity of proximate rank}
\label{sec:complexity}

\Cref{algo:greedy-prank-bound} does \emph{not} compute the proximate rank
in all cases---merely an upper bound.
In this section, we draw on computational complexity theory to show that this
suboptimality is fundamental.
It suffices to study the following decision problem, formalising the task of
bounding the proximate rank, or equivalently, detecting nearby low-rank
parameters.
The problem of computing the proximate rank of a given parameter is at least
as hard as this decision problem.

\begin{problem}[\PAR]
    \label{prob:par}
    \emph{Bounding proximate rank} (\PAR) is a decision problem.
    Each instance comprises
        a number of hidden units $h \in \Natz$,
        a parameter $w\in\W_h$,
        a uniform radius $\eps\in\PosReals$,
    and a maximum rank $r\in\Natz$.
    Affirmative instances are those for which
        $\prank\eps(w) \leq r$.
\end{problem}

Our main result is that \probPAR{} is \NP-complete (\cref{thm:par-npc}).
\NP-complete decision problems have no polynomial-time exact solutions unless
$\Poly = \NP$, though in many cases, good approximations are achievable in
practice (see \cref{sec:discussion}).

To show that \probPAR{} is \NP-complete, we must show that \probPAR{} is in
\NP{} and that all problems in \NP{} are reducible to \probPAR{}.
For the latter, since reducibility is transitive, it suffices to reduce a
single known \NP-hard problem to \probPAR{}.

Intuitively, there are two opportunities for the bound from
\cref{algo:greedy-prank-bound} to be suboptimal.
\begin{enumerate}
    \item
        \Cref{algo:greedy-approx-partition} does not in general find an
        optimal approximate partition. If there exists a more efficient
        approximate partition, this indicates a nearby parameter with
        lower rank.
    \item
        Moreover, even given an optimal approximate partition, one must
        additionally account for groups of units that have
            \emph{low outgoing weight after merging},
        indicating nearby parameters for which the corresponding merged units
        can be eliminated, leading to even lower rank.
\end{enumerate}
Each of~(1) and~(2) represent an opportunity to embed a known hard
computational problem into an instance of \probPAR{}.
We exploit~(1) and give a reduction from the following \NP-complete decision
problem, \probUPC, capturing the task of finding good approximate
partitions.\footnote{%
    \Cref{apx:unbiased} gives an alternative proof based on~(2) by
    reduction from subset sum.
}

\begin{problem}[\UPC]
    \label{prob:upc}
    \emph{Uniform point cover} (\UPC) is a decision problem.
    Each instance comprises a collection of $h \in \Natz$ source points
        $\subrange1hx \in \Reals^2$,
    a uniform radius $\eps \in \PosReals$,
    and a number of covering points $r \in \Natz$.
    Define an \emph{$(r,\eps)$-cover} of the source points as a collection of
    $r$ \emph{covering points} $\subrange1ry \in \Reals^2$
    such that the uniform distance between each source point and its nearest
    covering point is at most $\eps$ (that is,
    $
        \forall i \in \set{1, \ldots, h},
            \exists j \in \set{1, \ldots, r},
                \cdist{x_i}{y_j} \leq \eps
    $).
    Affirmative instances of \probUPC{} are those for which there exists an
        $(r,\eps)$-cover of $\subrange1hx$.
\end{problem}

\begin{theorem}
    \label{thm:upc-npc}
    \probUPC{} is \NP-complete.\footnotemark{}
\end{theorem}

\begin{figure}[!h]
    \centering
    \includegraphics[width=\textwidth]{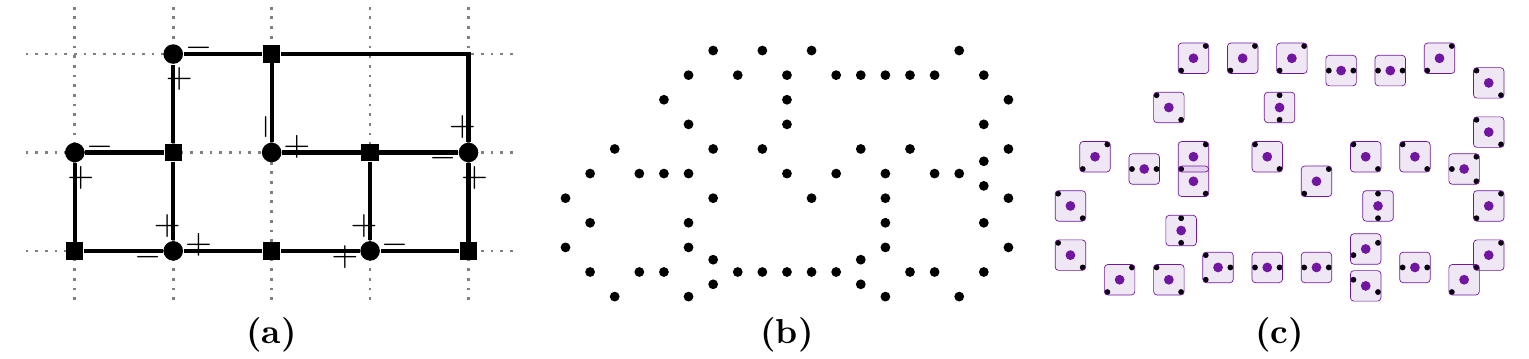}
    \caption{\label{fig:light-overview}%
        Example of reduction from Boolean satisfiability to \probUPC.
        \textbf{(a)}~Grid layout of an incidence graph for a
        satisfiable Boolean formula 
            (circles: variables, squares: clauses, edges: $\pm$ literals).
        \textbf{(b)}~A corresponding set of $h=68$ source points.
        \textbf{(c)}~A $(34,\eps)$-cover of the source points.
    }
\end{figure}

\begin{proofsketch}[Full proof in \cref{apx:complexity}]
    The proof is by reduction from Boolean satisfiability, a canonical
    \NP-complete decision problem, and involves constructing a set of source
    points tracing out a circuit representation of a Boolean formula, such that
    a sufficiently small cover exists if and only if the formula is satisfiable
    (illustrated in \cref{fig:light-overview}, full details in
    \cref{apx:complexity}).
\end{proofsketch}

\footnotetext{%
    An equivalent problem was shown to be \NP-hard by
        \citet{Megiddo+Supowit1984}.
    We discuss this equivalent problem and other related problems from the
    computational complexity literature in \cref{apx:upc}.
}

\newpage

Armed with \cref{thm:upc-npc} we can proceed to the main result of this
section.

\begin{theorem}
    \label{thm:par-npc}
    \probPAR{} is \NP-complete.
\end{theorem}

\begin{proofsketch}[Full proof below]
    The reduction from \probUPC{} constitutes the bulk of the proof, and
    proceeds as follows.
    Given an instance of \probUPC, allocate one hidden unit per source point,
    and construct a parameter using the source point coordinates as incoming
    weights and biases.
    Actually, first translate the source points well into the positive
    quadrant. Likewise, set the outgoing weights to a sufficiently positive
    value.
    \Cref{fig:prank-reduction} illustrates this construction.

    If there is a small cover of the source points then the covering points
    reveal the weights of nearby low-rank parameter.
    Conversely, if the resulting parameter has bounded proximate rank, it
    must be because of approximate merging (no other reducibility conditions
    approximately hold because all weights are sufficiently positive),
    so the nearby low-rank parameter reveals a small cover.
\end{proofsketch}

\begin{myproof}
($\UPC \reducto \PAR$, reduction):
    Consider an instance of \probUPC{} with
        $h, r \in \Natz$, $\eps \in \PosReals$, and
        source points $x_1, \ldots, x_h \in \Reals^2$.
    In linear time construct a \PAR{} instance with
        $h$ hidden units, uniform radius $\eps$, maximum rank $r$,
    and parameter $w \in \W_h$ as follows
        (see also \cref{fig:prank-reduction}).
    \begin{enumerate}
        \item
            Define
                $
                    x_{\min} = \left(
                        \min_{i=1}^h x_{i,1},
                        \min_{i=1}^h x_{i,2}
                    \right)
                    \in \Reals^2
                $,
            containing the minimum first and second coordinates among all
            source points (minimising over each dimension independently).
        \item
            Define a translation
                $T : \Reals^2 \to \Reals^2$
            such that
                $T(x) = x - x_{\min} + (2\eps, 2\eps)$.
        \item
            Translate the source points
                $\subrange1hx$
            to 
                $\subrange1h{x'}$
            where
                $x'_i = T(x_i)$.
            Note (for later) that all components of the translated
            source points are at least $2\eps$ by step (1).
        \item
            Construct the neural network parameter
                $
                    w = (
                        2\eps, x'_{1,1}, x'_{1,2},
                        \ldots,
                        2\eps, x'_{h,1}, x'_{h,2},
                        0
                    ) \in \W_h
                $.
            In other words, for $i = \range1h$, set
                $a_i = 2\eps$, 
                $b_i = x'_{i,1}$, and
                $c_i = x'_{i,2}$;
            and set $d=0$.
    \end{enumerate}

\begin{figure}[h!]
    \centering
    \includegraphics[]{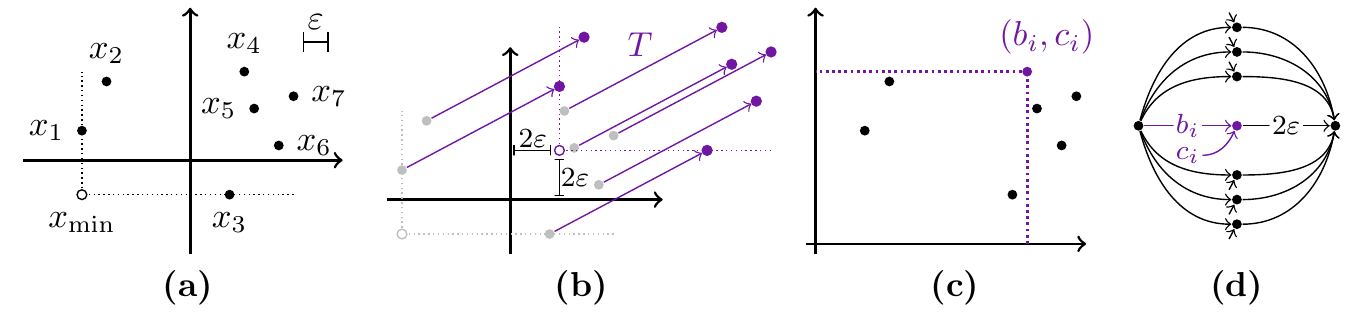}
    \caption{\label{fig:prank-reduction}%
        Illustrative example of the parameter construction.
        \textbf{(a)}~A set of source points $\subrange17x$.
        \textbf{(b)}~Transformation $T$ translates all points into
            the positive quadrant by a margin of $2\eps$.
        \textbf{(c,d)}~The coordinates of the transformed points become the
            incoming weights and biases of the parameter.
    }
\end{figure}

($\UPC \reducto \PAR$, equivalence):
    It remains to show that the constructed instance of \PAR{} is
    affirmative if and only if the given instance of \UPC{} is affirmative,
    that is,
        there exists an $(r,\eps)$-cover of the source points
    if and only if
        the constructed parameter has $\prank\eps(w) \leq r$.

($\limp$):
    If there is a small cover of the source points, then the hidden units can
    be perturbed so that they match up with the (translated) covering points.
    Since there are few covering points, many units can now be merged,
    so the original parameter has low proximate rank.

    Formally, suppose there exists an $(r,\eps)$-cover $\subrange1ry$.
    Define $\rho : \setrange1h \to \setrange1r$ such that
        the nearest covering point to each source point $x_i$ is
        $y_{\rho(i)}$
    (breaking ties arbitrarily).
    Then for $j=\range1r$, define $y'_j = T(y_j)$ where $T$ is the
    translation defined in step~(2) of the construction.
    Finally, define a parameter
        $
            w^\star
            = (
                2\eps, y'_{\rho(1),1}, y'_{\rho(1),2},
                \ldots,
                2\eps, y'_{\rho(h),1}, y'_{\rho(h),2},
                0
            ) \in \W_h
        $
    (in other words, for $i=\range1h$,
        $a_i^\star = 2\eps$,
        $b_i^\star = y'_{\rho(i),1}$, and
        $c_i^\star = y'_{\rho(i),2}$; and
        $d^\star = 0$).

    Then $\rank(w^\star) \leq r$, since there are at most $r$ distinct
    incoming weight and bias pairs (namely $\subrange1r{y'}$).
    Moreover, $\cdist{w}{w^\star} \leq \eps$, since both parameters have
    the same output bias and outgoing weights, and, by the defining property
    of the cover, for $i=\range1h$,
    \begin{equation*}
        \cdist{(b_i,c_i)}{(b_i^\star,c_i^\star)}
        = \cdist{x'_i}{y'_{\rho(i)}}
        = \cdist{T(x_i)}{T(y_{\rho(i)})}
        = \cdist{x_i}{y_{\rho(i)}}
        \leq \eps.
    \end{equation*}
    Therefore $\prank\eps(w) \leq \rank(w^\star) \leq r$.

($\Leftarrow$):
    Conversely, since all of the weights and biases are at least $2\eps$, any
    nearby low-rank parameter implies the approximate mergeability of some
    units.
    Therefore, if the parameter has low proximate rank, there
    is a small cover of the translated points, and, in turn, of the original
    points.

    Formally, suppose
        $\prank\eps(w) \leq r$,
    with $w^\star \in \cnbhd\eps{w}$ such that
        $\rank(w^\star) = r^\star \leq r$.
    In general, the only ways that $w^\star$ could have reduced rank
    compared to $w$ are the following (cf.~\cref{algo:lossless-compression}):
    \begin{enumerate}
        \item
            Some incoming weight $b_i$ could be perturbed to zero,
            allowing its unit to be eliminated.
        \item
            Two units $i, j$ with $(b_i, c_i)$ and $(b_j, c_j)$ within
            $2\eps$ could be perturbed to have identical incoming weight and
            bias, allowing them to be merged.
        \item
            Two units $i, j$ with $(b_i, c_i)$ and $-(b_j, c_j)$ within
            $2\eps$ could be perturbed to have identically negative weight
            and bias, again allowing them to be merged.
        \item
            Some group of $m \geq 1$ units, merged through the above options,
            with total outgoing weight within $m\eps$ of zero,
            could have their outgoing weights perturbed to make the total
            zero.
    \end{enumerate}
    By construction, all $a_i, b_i, c_i \geq 2\eps > 0$, immediately ruling
    out~(1) and~(3). Option~(4) is also ruled out because any such total
    outgoing weight is $2m\eps > m\eps$. This leaves option~(2) alone
    responsible. Thus, there are exactly $r^\star$ distinct incoming weight
    and bias pairs among the units of $w^\star$.
    Denote these pairs $\subrange1{r^\star}{y'}$---they constitute an
    $(r^\star,\eps)$-cover of the incoming weight and bias vectors of $w$,
        $\subrange1h{x'}$
    (as $w^\star \in \cnbhd\eps{w}$).
    Finally, invert $T$ to produce an $(r^\star,\eps)$-cover of
    $\subrange1hx$, and add $r-r^\star$ arbitrary covering points to extend
    this to the desired $(r,\eps)$-cover.

($\PAR \in \NP$):
    We must show that an affirmative instance of \PAR{} can be verified in
    polynomial time, given a certificate.
    Consider an instance $h, r \in \Natz$, $\eps \in \PosReals$, and
        $w = (\abcrange, d) \in \W_h$.
    Use as a certificate a partition\footnotemark{}
        $\subrange1J\Pi$
    of $\Set{i\in\setrange1h}{\cnorm{b_i} > \eps}$,
    such that (1)~for each $\Pi_j$, for each $i, k \in \Pi_j$,
        $\cdist{\sign(b_i)\cdot(b_i,c_i)}{\sign(b_k)\cdot(b_k,c_k)}\leq2\eps$;
    and (2)~at most $r$ of the $\Pi_j$ satisfy
        $\sum_{i\in\Pi_j} \sign(b_i) \cdot a_i>\eps\cdot\cardinality{\Pi_j}$.
    The validity of such a certificate can be verified in polynomial time by
    checking each of these conditions directly.
    
    \footnotetext{\label{fn:certification}%
        It would seem simpler to use a nearby low-rank parameter as the
        certificate, since by definition one exists exactly in affirmative
        cases.
        However, we would separately have to establish that such a parameter
        has polynomial description length, since an arbitrary real parameter
        does not.
        Using a partition of hidden units implicitly establishes this by
        describing the reducible parameter in terms of the existing
        parameter.
    }
    
    It remains to show that such a certificate exists if and only if the
    instance is affirmative.
    If $\prank\eps(w) \leq r$, then there exists a parameter
        $w^\star \in \cnbhd\eps{w}$
    with $\rank(w^\star) \leq r$.
    The partition computed from Stage~2 of $\CALL{Compress}(w^\star)$
    satisfies the required properties for $w \in \cnbhd\eps{w^\star}$.
     
    Conversely, given such a partition, for each $\Pi_j$, define
    $v_j \in \Reals^2$ as the centroid of the bounding rectangle of the
    set of points $\Set{\sign(b_i)\cdot(b_i,c_i)}{i\in\Pi_j}$,
    that is,
        \begin{equation*}
            v_j = \frac12\left(\ 
                \max_{i \in \Pi_j} \sign(b_i) \cdot b_i
                +
                \min_{i \in \Pi_j} \sign(b_i) \cdot b_i
                ,\quad
                \max_{i \in \Pi_j} \sign(b_i) \cdot c_i
                +
                \min_{i \in \Pi_j} \sign(b_i) \cdot c_i
            \ \right)\!.
        \end{equation*}
    All of the points within these bounding rectangles are at most uniform
    distance $\eps$ from their centroids.
    To construct a nearby low-rank parameter, follow the proof of
        \cref{thm:algo:greedy-prank-bound}
    using $\subrange1J\Pi$ and $\subrange1Jv$ in place of their namesakes
    from \cref{algo:greedy-prank-bound,algo:greedy-approx-partition}.
    Thus $\prank\eps(w) \leq r$.
\end{myproof}

\section{Discussion}
\label{sec:discussion}

We have developed an algorithmic framework for lossless compressibility in
single-hidden-layer hyperbolic tangent networks.
The \emph{rank} measures a parameter's lossless compressibility.
\Cref{sec:rank} offers efficient algorithms for performing optimal lossless
compression and computing the rank.
The \emph{proximate rank} measures proximity to low-rank parameters.
\Cref{sec:prank} offers an efficient algorithm approximately bounding the
proximate rank.
\Cref{sec:complexity} shows that optimally bounding the proximate rank is
\NP-complete, connecting proximate rank to well-studied problems in computer
science.

While losslessly compressible parameters are themselves atypical, being
\emph{near} a losslessly compressible parameter is a sufficient condition for
a parameter to be \emph{approximately} compressible, since similar parameters
implement similar functions.
Therefore, studying these parameters could give a new lens on the phenomenon
of approximate compressibility in deep learning.

Our results lay a foundation for future theoretical and empirical work
investigating this perspective on approximate compressibility.
We discuss potential limitations of this approach and important questions
for future work below.

\paragraph{Does deep learning find solutions with low proximate rank in
practice?}

While low proximate rank implies approximate compressibility, the converse is
false.
In particular, approximate compression approaches often require the
implemented function to be similar only for certain inputs.
Moreover, there are typically ways of implementing similar functions with
dissimilar parameters \citep{Petersen+2021}.
It is an open question to what extent the approximate compressibility
observed in deep learning practice coincides with proximity to losslessly
compressible parameters.

There is evidence that lossless compressibility is relevant in at least some
cases.
While investigating the structure of learned neural networks,
\citet{Casper+2021} found many units with weak or correlated outputs, and
found that these units could be removed without a large effect on
performance, using elimination and merging operations bearing a striking
resemblance to Sussmann's reducibility operations (\cref{sec:prelims}).
Deeply studying this phenomenon and other similar empirical questions
suggested by the theory of lossless compressibility is an important direction
for future work.

\paragraph{Can low proximate rank parameters be detected in practice?}

\Cref{sec:complexity} suggests that detecting low proximate rank is
computationally intractable in general.
However, this should not be overstated.
This is a worst-case analysis---\cref{thm:par-npc,thm:upc-npc} essentially
construct pathological neural network parameters poised precisely between
nearby low-rank regions such that finding the \emph{most} compressible nearby
parameter involves solving an instance of Boolean satisfiability.

Instances of low proximate rank encountered in practice may be easy to
detect. This would be typical---solving and approximately solving
\NP-complete problems is a routine operation in many practical settings.
Take the example of $k$-means clustering, an \NP-complete problem for which a
simple algorithm achieves good results in many practical instances
    \citep{Daniely+2012}.
Boolean satisfiability itself also has both easy and hard instances
    \citep{Cheeseman+1991}.
Moreover, suboptimal bounds on proximate rank are still useful as a one-sided
test of approximate compressibility.

Rather, by establishing that this problem is \NP-complete, we have
established precise connections to a range of well-studied problems
    (\cref{apx:upc}),
opening the door to a wealth of literature on approximation approaches that
can be tried
    \citep[cf.][\S6]{Garey+Johnson1979}.
\Cref{algo:greedy-prank-bound} provides a naive approximation serving as a
baseline with room for improvement in future work.

\paragraph{What additional opportunities for lossless compression arise in
more complex architectures?}

The most salient limitation of this work is the setting of
single-hidden-layer hyperbolic tangent networks. This setting determines some
of the details of our algorithms, theorem statements, and proofs.
While this architecture is not immediately relevant to modern deep learning,
our results are a meaningful first step towards an analysis of lossless
compressibility in larger architectures.

Much of our analysis applies generically to individual feed-forward layers in
multi-layer architectures with any nonlinearity.
At the core of our analysis are redundancies arising from zero,
constant, or proportional units (cf.~reducibility conditions
    \ref{item:reducibility:1}--\ref{item:reducibility:3}).
These conditions arise from the computational structure of any neuron or
layer of neurons, independently of the hyperbolic tangent nonlinearity.
Only condition \ref{item:reducibility:4} is non-generic, but neither is it
fundamental to our analysis.
Thus studying this concrete case has allowed us to analysed three fundamental
forms of redundancy that can arise in any neural network, and to outline ways
to exploit them for (approximate) lossless compression.

In more complex architectures, these compression opportunities will remain,
but there will be \emph{additional} forms of redundancy that arise, for
example due to the affine symmetries of the chosen nonlinearity or due to
interactions between layers in a multi-layer network.
Such additional forms of redundancy have not been catalogued.
The lossless compression framework offers a principled objective to aid in
identifying them and responding to them.

We call for future work to ask the question, what does optimal lossless
compression look like in more complex architectures?
Such work can use our algorithms and analysis for the single-hidden-layer
case as a starting point, extending these to exploit novel redundancies in
addition to the fundamental redundancies we have identified.
In turn, such work can suggest new empirical questions arising from this new
perspective on approximate compressibility in learned neural networks.


\newpage
\begin{ack}

The contributions in this paper also appear in MFR's minor thesis
    \citep[\S4 and~\S6]{Farrugia2022}.
MFR received financial support from the Melbourne School of Engineering
Foundation Scholarship and the Long-Term Future Fund while completing
this research.
MFR would like to thank
    Daniel Murfet,
    Harald S\o{}ndergaard,
    and
    Rohan Hitchcock
for providing helpful feedback during this research and during the
preparation of this manuscript.
\end{ack}


\bibliographystyle{far}
\bibliography{main}

\begin{thebibliography}{59}
\providecommand{\natexlab}[1]{#1}
\providecommand{\doi}[1]{\href{https://doi.org/#1}{Crossref}}
\providecommand{\viaurl}[2][web]{\href{#2}{#1}}
\providecommand{\eprint}[2]{\href{https://arxiv.org/abs/#1}{arXiv:#1} {\small
  [#2]}}

\bibitem[{Albertini et~al.(1993)Albertini, Sontag, and
  Maillot}]{Albertini+1993}
Francesca Albertini, Eduardo~D. Sontag, and Vincent Maillot.
\newblock Uniqueness of weights for neural networks.
\newblock In \emph{Artificial Neural Networks for Speech and Vision}, pages
  113--125. Chapman \& Hall, London, 1993.
\newblock Proceedings of a workshop held at Rutgers University in 1992. Access
  via \viaurl[Eduardo D. Sontag]{http://www.sontaglab.org/FTPDIR/92caip.pdf}.

\bibitem[{Amari et~al.(2006)Amari, Park, and Ozeki}]{Amari+2006}
Shun{-}ichi Amari, Hyeyoung Park, and Tomoko Ozeki.
\newblock Singularities affect dynamics of learning in neuromanifolds.
\newblock \emph{Neural Computation}, 18(5):1007--1065, 2006.
\newblock Access via \doi{10.1162/neco.2006.18.5.1007}.

\bibitem[{Amari et~al.(2018)Amari, Ozeki, Karakida, Yoshida, and
  Okada}]{Amari+2018}
Shun{-}ichi Amari, Tomoko Ozeki, Ryo Karakida, Yuki Yoshida, and Masato Okada.
\newblock Dynamics of learning in {MLP}: Natural gradient and singularity
  revisited.
\newblock \emph{Neural Computation}, 30(1):1--33, 2018.
\newblock Access via \doi{10.1162/neco_a_01029}.

\bibitem[{Aytekin et~al.(2019)Aytekin, Cricri, and Aksu}]{Aytekin+2019}
Caglar Aytekin, Francesco Cricri, and Emre Aksu.
\newblock Compressibility loss for neural network weights.
\newblock 2019.
\newblock Preprint \eprint{1905.01044}{cs.LG}.

\bibitem[{Berman et~al.(2003)Berman, Scott, and Karpinski}]{Berman+2003}
Piotr Berman, Alex~D. Scott, and Marek Karpinski.
\newblock Approximation hardness and satisfiability of bounded occurrence
  instances of {SAT}.
\newblock Technical Report IHES/M/03/25, Institut des Hautes {\'E}tudes
  Scientifiques [Institute of Advanced Scientific Studies], 2003.
\newblock Access via \viaurl[CERN]{https://cds.cern.ch/record/630836/}.

\bibitem[{Bucilu\v{a} et~al.(2006)Bucilu\v{a}, Caruana, and
  Niculescu-Mizil}]{Bucilua+2006}
Cristian Bucilu\v{a}, Rich Caruana, and Alexandru Niculescu-Mizil.
\newblock Model compression.
\newblock In \emph{Proceedings of the 12th ACM SIGKDD International Conference
  on Knowledge Discovery and Data Mining}, pages 535--541. ACM, 2006.
\newblock Access via \doi{10.1145/1150402.1150464}.

\bibitem[{Casper et~al.(2021)Casper, Boix, D'Amario, Guo, Schrimpf, Vinken, and
  Kreiman}]{Casper+2021}
Stephen Casper, Xavier Boix, Vanessa D'Amario, Ling Guo, Martin Schrimpf,
  Kasper Vinken, and Gabriel Kreiman.
\newblock Frivolous units: Wider networks are not really that wide.
\newblock In \emph{Proceedings of the Thirty-Fifth AAAI Conference on
  Artificial Intelligence}, volume~8, pages 6921--6929. AAAI Press, 2021.
\newblock Access via \doi{10.1609/aaai.v35i8.16853}.

\bibitem[{Cerioli et~al.(2004)Cerioli, Faria, Ferreira, and
  Protti}]{Cerioli+2004}
M{\'a}rcia~R. Cerioli, Luerbio Faria, Talita~O. Ferreira, and F{\'a}bio Protti.
\newblock On minimum clique partition and maximum independent set on unit disk
  graphs and penny graphs: Complexity and approximation.
\newblock \emph{Electronic Notes in Discrete Mathematics}, 18:73--79, 2004.
\newblock Access via \doi{10.1016/j.endm.2004.06.012}.

\bibitem[{Cerioli et~al.(2011)Cerioli, Faria, Ferreira, and
  Protti}]{Cerioli+2011}
M{\'a}rcia~R. Cerioli, Luerbio Faria, Talita~O. Ferreira, and F{\'a}bio Protti.
\newblock A note on maximum independent sets and minimum clique partitions in
  unit disk graphs and penny graphs: Complexity and approximation.
\newblock \emph{RAIRO: Theoretical Informatics and Applications},
  45(3):331--346, 2011.
\newblock Access via \doi{10.1051/ita/2011106}.

\bibitem[{Cheeseman et~al.(1991)Cheeseman, Kanefsky, and
  Taylor}]{Cheeseman+1991}
Peter Cheeseman, Bob Kanefsky, and William~M Taylor.
\newblock Where the really hard problems are.
\newblock In \emph{Proceedings of the Twelfth International Joint Conference on
  Artificial Intelligence}, volume~1, pages 331--337. 1991.

\bibitem[{Chen et~al.(1993)Chen, Lu, and Hecht-Nielsen}]{Chen+1993}
An~Mei Chen, Haw{-}minn Lu, and Robert Hecht-Nielsen.
\newblock On the geometry of feedforward neural network error surfaces.
\newblock \emph{Neural Computation}, 5(6):910--927, 1993.
\newblock Access via \doi{10.1162/neco.1993.5.6.910}.

\bibitem[{Cheng et~al.(2018)Cheng, Wang, Zhou, and Zhang}]{Cheng+2018}
Yu~Cheng, Duo Wang, Pan Zhou, and Tao Zhang.
\newblock Model compression and acceleration for deep neural networks: The
  principles, progress, and challenges.
\newblock \emph{IEEE Signal Processing Magazine}, 35(1):126--136, 2018.
\newblock Access via \doi{10.1109/MSP.2017.2765695}.

\bibitem[{Cheng et~al.(2020)Cheng, Wang, Zhou, and Zhang}]{Cheng+2020}
Yu~Cheng, Duo Wang, Pan Zhou, and Tao Zhang.
\newblock A survey of model compression and acceleration for deep neural
  networks.
\newblock 2020.
\newblock Preprint \eprint{1710.09282v9}{cs.LG}.
\newblock Updated version of \citet{Cheng+2018}.

\bibitem[{Choudhary et~al.(2020)Choudhary, Mishra, Goswami, and
  Sarangapani}]{Choudhary+2020}
Tejalal Choudhary, Vipul Mishra, Anurag Goswami, and Jagannathan Sarangapani.
\newblock A comprehensive survey on model compression and acceleration.
\newblock \emph{Artificial Intelligence Review}, 53(7):5113--5155, 2020.
\newblock Access via \doi{10.1007/s10462-020-09816-7}.

\bibitem[{Clark et~al.(1990)Clark, Colbourn, and Johnson}]{Clark+1990}
Brent~N. Clark, Charles~J. Colbourn, and David~S. Johnson.
\newblock Unit disk graphs.
\newblock \emph{Discrete Mathematics}, 86(1-3):165--177, 1990.
\newblock Access via \doi{10.1016/0012-365X(90)90358-O}.

\bibitem[{Cook(1971)}]{Cook1971}
Stephen~A. Cook.
\newblock The complexity of theorem-proving procedures.
\newblock In \emph{Proceedings of the Third Annual ACM Symposium on Theory of
  Computing}, pages 151--158. ACM, 1971.
\newblock Access via \doi{10.1145/800157.805047}.

\bibitem[{Cousseau et~al.(2008)Cousseau, Ozeki, and Amari}]{Cousseau+2008}
Florent Cousseau, Tomoko Ozeki, and Shun{-}ichi Amari.
\newblock Dynamics of learning in multilayer perceptrons near singularities.
\newblock \emph{IEEE Transactions on Neural Networks}, 19(8):1313--1328, 2008.
\newblock Access via \doi{10.1109/TNN.2008.2000391}.

\bibitem[{Daniely et~al.(2012)Daniely, Linial, and Saks}]{Daniely+2012}
Amit Daniely, Nati Linial, and Michael Saks.
\newblock Clustering is difficult only when it does not matter.
\newblock 2012.
\newblock Preprint \eprint{1205.4891}{cs.LG}.

\bibitem[{Farrugia-Roberts(2022)}]{Farrugia2022}
Matthew Farrugia-Roberts.
\newblock \emph{Structural Degeneracy in Neural Networks}.
\newblock Master's thesis, School of Computing and Information Systems, The
  University of Melbourne, 2022.
\newblock Access via \viaurl[Matthew
  Farrugia-Roberts]{https://far.in.net/mthesis}.

\bibitem[{Farrugia-Roberts(2023)}]{Farrugia2023}
Matthew Farrugia-Roberts.
\newblock Functional equivalence and path connectivity of reducible hyperbolic
  tangent networks.
\newblock In \emph{Advances in Neural Information Processing Systems 36}, pages
  79502--79517. Curran Associates, 2023.
\newblock Access via
  \viaurl[NeurIPS]{https://proceedings.neurips.cc/paper_files/paper/2023/hash/fb64a43508e0cfe53ee6179ff31ea900-Abstract-Conference.html}.

\bibitem[{Fefferman(1994)}]{Fefferman1994}
Charles Fefferman.
\newblock Reconstructing a neural net from its output.
\newblock \emph{Revista Matem{\'a}tica Iberoamericana}, 10(3):507--555, 1994.
\newblock Access via \doi{10.4171/RMI/160}.

\bibitem[{Fefferman and Markel(1993)}]{Fefferman+Markel1993}
Charles Fefferman and Scott Markel.
\newblock Recovering a feed-forward net from its output.
\newblock In \emph{Advances in Neural Information Processing Systems 6}, pages
  335--342. Morgan Kaufmann, 1993.
\newblock Access via
  \viaurl[NeurIPS]{https://proceedings.neurips.cc/paper/1993/hash/e49b8b4053df9505e1f48c3a701c0682-Abstract.html}.

\bibitem[{Fukumizu(1996)}]{Fukumizu1996}
Kenji Fukumizu.
\newblock A regularity condition of the information matrix of a multilayer
  perceptron network.
\newblock \emph{Neural Networks}, 9(5):871--879, 1996.
\newblock Access via \doi{10.1016/0893-6080(95)00119-0}.

\bibitem[{Fukumizu and Amari(2000)}]{Fukumizu+Amari2000}
Kenji Fukumizu and Shun{-}ichi Amari.
\newblock Local minima and plateaus in hierarchical structures of multilayer
  perceptrons.
\newblock \emph{Neural Networks}, 13(3):317--327, 2000.
\newblock Access via \doi{10.1016/S0893-6080(00)00009-5}.

\bibitem[{Fukumizu et~al.(2019)Fukumizu, Yamaguchi, Mototake, and
  Tanaka}]{Fukumizu+2019}
Kenji Fukumizu, Shoichiro Yamaguchi, Yoh-ichi Mototake, and Mirai Tanaka.
\newblock Semi-flat minima and saddle points by embedding neural networks to
  overparameterization.
\newblock In \emph{Advances in Neural Information Processing Systems 32}, pages
  13868--13876. Curran Associates, 2019.
\newblock Access via
  \viaurl[NeurIPS]{https://proceedings.neurips.cc/paper/2019/hash/a4ee59dd868ba016ed2de90d330acb6a-Abstract.html}.

\bibitem[{Garey and Johnson(1979)}]{Garey+Johnson1979}
Michael~R. Garey and David~S. Johnson.
\newblock \emph{Computers and Intractability: A Guide to the Theory of
  NP-Completeness}.
\newblock W. H. Freeman and Company, 1979.

\bibitem[{Hakimi(1964)}]{Hakimi1964}
S.~L. Hakimi.
\newblock Optimum locations of switching centers and the absolute centers and
  medians of a graph.
\newblock \emph{Operations Research}, 12(3):450--459, 1964.
\newblock Access via \doi{10.1287/opre.12.3.450}.

\bibitem[{Hakimi(1965)}]{Hakimi1965}
S.~L. Hakimi.
\newblock Optimum distribution of switching centers in a communication network
  and some related graph theoretic problems.
\newblock \emph{Operations Research}, 13(3):462--475, 1965.
\newblock Access via \doi{10.1287/opre.13.3.462}.

\bibitem[{Hinton and van Camp(1993)}]{Hinton+vanCamp1993}
Geoffrey~E. Hinton and Drew van Camp.
\newblock Keeping the neural networks simple by minimizing the description
  length of the weights.
\newblock In \emph{Proceedings of the Sixth Annual Conference on Computational
  Learning Theory}, pages 5--13. ACM, 1993.
\newblock Access via \doi{10.1145/168304.168306}.

\bibitem[{Hinton et~al.(2014)Hinton, Vinyals, and Dean}]{Hinton+2015}
Geoffrey~E. Hinton, Oriol Vinyals, and Jeff Dean.
\newblock Distilling the knowledge in a neural network.
\newblock Presented at Twenty-eighth Conference on Neural Information
  Processing Systems, Deep Learning workshop, 2014.
\newblock Preprint \eprint{1503.02531}{stat.ML}.

\bibitem[{Jansen and M{\"u}ller(1995)}]{Jansen+Muller1995}
Klaus Jansen and Haiko M{\"u}ller.
\newblock The minimum broadcast time problem for several processor networks.
\newblock \emph{Theoretical Computer Science}, 147(1-2):69--85, 1995.
\newblock Access via \doi{10.1016/0304-3975(94)00230-G}.

\bibitem[{Kariv and Hakimi(1979)}]{Kariv+Hakimi1979}
O.~Kariv and S.~L. Hakimi.
\newblock An algorithmic approach to network location problems. {I:} the
  $p$-centers.
\newblock \emph{SIAM Journal on Applied Mathematics}, 37(3):513--538, 1979.
\newblock Access via \doi{10.1137/0137040}.

\bibitem[{Karp(1972)}]{Karp1972}
Richard~M. Karp.
\newblock Reducibility among combinatorial problems.
\newblock In \emph{Complexity of Computer Computations}, pages 85--103.
  Springer, 1972.
\newblock Access via \doi{10.1007/978-1-4684-2001-2_9}.

\bibitem[{K\r{u}rkov\'{a} and Kainen(1994)}]{Kurkova+Kainen1994}
V\v{e}ra K\r{u}rkov\'{a} and Paul~C. Kainen.
\newblock Functionally equivalent feedforward neural networks.
\newblock \emph{Neural Computation}, 6(3):543--558, 1994.
\newblock Access via \doi{10.1162/neco.1994.6.3.543}.

\bibitem[{Levin(1973)}]{Levin1973}
Leonid~A. Levin.
\newblock Universal sequential search problems.
\newblock \emph{Problemy Peredachi Informatsii [Problems of Information
  Transmission]}, 9(3):115--116, 1973.
\newblock In Russian. Translated into English in \citet{Trakhtenbrot1984}.

\bibitem[{Lichtenstein(1982)}]{Lichtenstein1982}
David Lichtenstein.
\newblock Planar formulae and their uses.
\newblock \emph{SIAM Journal on Computing}, 11(2):329--343, 1982.
\newblock Access via \doi{10.1137/0211025}.

\bibitem[{Liu et~al.(1998)Liu, Morgana, and Simeone}]{Liu+1998}
Yanpei Liu, Aurora Morgana, and Bruno Simeone.
\newblock A linear algorithm for 2-bend embeddings of planar graphs in the
  two-dimensional grid.
\newblock \emph{Discrete Applied Mathematics}, 81(1-3):69--91, 1998.
\newblock Access via \doi{10.1016/S0166-218X(97)00076-0}.

\bibitem[{Mahajan et~al.(2012)Mahajan, Nimbhorkar, and
  Varadarajan}]{Mahajan+2012}
Meena Mahajan, Prajakta Nimbhorkar, and Kasturi Varadarajan.
\newblock The planar $k$-means problem is {NP-hard}.
\newblock \emph{Theoretical Computer Science}, 442:13--21, 2012.
\newblock Access via \doi{10.1016/j.tcs.2010.05.034}.

\bibitem[{Megiddo and Supowit(1984)}]{Megiddo+Supowit1984}
Nimrod Megiddo and Kenneth~J. Supowit.
\newblock On the complexity of some common geometric location problems.
\newblock \emph{SIAM Journal on Computing}, 13(1):182--196, 1984.
\newblock Access via \doi{10.1137/0213014}.

\bibitem[{Petersen et~al.(2021)Petersen, Raslan, and
  Voigtlaender}]{Petersen+2021}
Philipp Petersen, Mones Raslan, and Felix Voigtlaender.
\newblock Topological properties of the set of functions generated by neural
  networks of fixed size.
\newblock \emph{Foundations of Computational Mathematics}, 21(2):375--444,
  2021.
\newblock Access via \doi{10.1007/s10208-020-09461-0}.

\bibitem[{Phuong and Lampert(2020)}]{Phuong+Lampert2020}
Mary Phuong and Christoph~H. Lampert.
\newblock Functional vs{.} parametric equivalence of {ReLU} networks.
\newblock In \emph{8th International Conference on Learning Representations}.
  OpenReview, 2020.
\newblock Access via
  \viaurl[OpenReview]{https://openreview.net/forum?id=Bylx-TNKvH}.

\bibitem[{Piziak and Odell(1999)}]{Piziak+Odell1999}
R.~Piziak and P.~L. Odell.
\newblock Full rank factorization of matrices.
\newblock \emph{Mathematics Magazine}, 72(3):193--201, 1999.
\newblock Access via \doi{10.1080/0025570X.1999.11996730}.

\bibitem[{Sanh et~al.(2019)Sanh, Debut, Chaumond, and Wolf}]{DistilBERT2020}
Victor Sanh, Lysandre Debut, Julien Chaumond, and Thomas Wolf.
\newblock {DistilBERT}, a distilled version of {BERT}: Smaller, faster, cheaper
  and lighter.
\newblock Presented at the Fifth Workshop on Energy Efficient Machine Learning
  and Cognitive Computing, 2019.
\newblock Preprint \eprint{1910.01108}{cs.CL}.

\bibitem[{Serra et~al.(2020)Serra, Kumar, and Ramalingam}]{Serra+2020}
Thiago Serra, Abhinav Kumar, and Srikumar Ramalingam.
\newblock Lossless compression of deep neural networks.
\newblock In \emph{Integration of Constraint Programming, Artificial
  Intelligence, and Operations Research: 17th International Conference, CPAIOR
  2020, Vienna, Austria, September 21--24, 2020, Proceedings}, pages 417--430.
  2020.
\newblock Access via \doi{10.1007/978-3-030-58942-4_27}.

\bibitem[{{\c{S}}im\c{s}ek et~al.(2021){\c{S}}im\c{s}ek, Ged, Jacot, Spadaro,
  Hongler, Gerstner, and Brea}]{Simsek+2021}
Berfin {\c{S}}im\c{s}ek, Fran{\c{c}}ois Ged, Arthur Jacot, Francesco Spadaro,
  Cl{\'e}ment Hongler, Wulfram Gerstner, and Johanni Brea.
\newblock Geometry of the loss landscape in overparameterized neural networks:
  Symmetries and invariances.
\newblock In \emph{Proceedings of the 38th International Conference on Machine
  Learning}, pages 9722--9732. PMLR, 2021.
\newblock Access via
  \viaurl[PMLR]{https://proceedings.mlr.press/v139/simsek21a.html}.

\bibitem[{Supowit(1981)}]{Supowit1981}
Kenneth~J. Supowit.
\newblock \emph{Topics in Computational Geometry}.
\newblock Ph.D. thesis, University of Illinois at Urbana-Champaign, 1981.
\newblock Access via
  \viaurl[ProQuest]{https://www.proquest.com/openview/318d326449b3dfbb7f23a0719877b241/1}.

\bibitem[{Sussmann(1992)}]{Sussmann1992}
H{\'e}ctor~J. Sussmann.
\newblock Uniqueness of the weights for minimal feedforward nets with a given
  input-output map.
\newblock \emph{Neural Networks}, 5(4):589--593, 1992.
\newblock Access via \doi{10.1016/S0893-6080(05)80037-1}.

\bibitem[{Suzuki et~al.(2020{\natexlab{a}})Suzuki, Abe, Murata, Horiuchi, Ito,
  Wachi, Hirai, Yukishima, and Nishimura}]{Suzuki+2020a}
Taiji Suzuki, Hiroshi Abe, Tomoya Murata, Shingo Horiuchi, Kotaro Ito, Tokuma
  Wachi, So~Hirai, Masatoshi Yukishima, and Tomoaki Nishimura.
\newblock Spectral pruning: Compressing deep neural networks via spectral
  analysis and its generalization error.
\newblock In \emph{Proceedings of the Twenty-Ninth International Joint
  Conference on Artificial Intelligence}, pages 2839--2846. IJCAI,
  2020{\natexlab{a}}.
\newblock Access via \doi{10.24963/ijcai.2020/393}.

\bibitem[{Suzuki et~al.(2020{\natexlab{b}})Suzuki, Abe, and
  Nishimura}]{Suzuki+2020b}
Taiji Suzuki, Hiroshi Abe, and Tomoaki Nishimura.
\newblock Compression based bound for non-compressed network: Unified
  generalization error analysis of large compressible deep neural network.
\newblock In \emph{8th International Conference on Learning Representations}.
  OpenReview, 2020{\natexlab{b}}.
\newblock Access via
  \viaurl[OpenReview]{https://openreview.net/forum?id=ByeGzlrKwH}.

\bibitem[{Tovey(1984)}]{Tovey1984}
Craig~A. Tovey.
\newblock A simplified {NP}-complete satisfiability problem.
\newblock \emph{Discrete Applied Mathematics}, 8(1):85--89, 1984.
\newblock Access via \doi{10.1016/0166-218X(84)90081-7}.

\bibitem[{Trakhtenbrot(1984)}]{Trakhtenbrot1984}
Boris~A. Trakhtenbrot.
\newblock A survey of {Russian} approaches to \foreign{perebor} (brute-force
  search) algorithms.
\newblock \emph{Annals of the History of Computing}, 6(4):384--400, 1984.
\newblock Access via \doi{10.1109/MAHC.1984.10036}.

\bibitem[{Valiant(1981)}]{Valiant1981}
Leslie~G. Valiant.
\newblock Universality considerations in {VLSI} circuits.
\newblock \emph{IEEE Transactions on Computers}, 100(2):135--140, 1981.
\newblock Access via \doi{10.1109/TC.1981.6312176}.

\bibitem[{Vla{\v{c}}i{\'c} and B{\"o}lcskei(2021)}]{Vlacic+Bolcskei2021}
Verner Vla{\v{c}}i{\'c} and Helmut B{\"o}lcskei.
\newblock Affine symmetries and neural network identifiability.
\newblock \emph{Advances in Mathematics}, 376:107485, 2021.
\newblock Access via \doi{10.1016/j.aim.2020.107485}.

\bibitem[{Vla{\v{c}}i{\'c} and B{\"o}lcskei(2022)}]{Vlacic+Bolcskei2022}
Verner Vla{\v{c}}i{\'c} and Helmut B{\"o}lcskei.
\newblock Neural network identifiability for a family of sigmoidal
  nonlinearities.
\newblock \emph{Constructive Approximation}, 55(1):173--224, 2022.
\newblock Access via \doi{10.1007/s00365-021-09544-3}.

\bibitem[{Watanabe(2009)}]{greybook}
Sumio Watanabe.
\newblock \emph{Algebraic Geometry and Statistical Learning Theory}.
\newblock Cambridge University Press, 2009.

\bibitem[{Wei et~al.(2008)Wei, Zhang, Cousseau, Ozeki, and Amari}]{Wei+2008}
Haikun Wei, Jun Zhang, Florent Cousseau, Tomoko Ozeki, and Shun{-}ichi Amari.
\newblock Dynamics of learning near singularities in layered networks.
\newblock \emph{Neural Computation}, 20(3):813--843, 2008.
\newblock Access via \doi{10.1162/neco.2007.12-06-414}.

\bibitem[{Wei et~al.(2023)Wei, Murfet, Gong, Li, Gell-Redman, and
  Quella}]{goodpaper}
Susan Wei, Daniel Murfet, Mingming Gong, Hui Li, Jesse Gell-Redman, and Thomas
  Quella.
\newblock Deep learning is singular, and that's good.
\newblock \emph{IEEE Transactions on Neural Networks and Learning Systems},
  34(12):10473--10486, 2023.
\newblock Access via \doi{10.1109/TNNLS.2022.3167409}.

\bibitem[{Zhang et~al.(2017)Zhang, Bengio, Hardt, Recht, and
  Vinyals}]{Zhang+2017}
Chiyuan Zhang, Samy Bengio, Moritz Hardt, Benjamin Recht, and Oriol Vinyals.
\newblock Understanding deep learning requires rethinking generalization.
\newblock In \emph{5th International Conference on Learning Representations}.
  OpenReview, 2017.
\newblock Access via
  \viaurl[OpenReview]{https://openreview.net/forum?id=Sy8gdB9xx}.

\bibitem[{Zhang et~al.(2021)Zhang, Bengio, Hardt, Recht, and
  Vinyals}]{Zhang+2021}
Chiyuan Zhang, Samy Bengio, Moritz Hardt, Benjamin Recht, and Oriol Vinyals.
\newblock Understanding deep learning (still) requires rethinking
  generalization.
\newblock \emph{Communications of the ACM}, 64(3):107--115, 2021.
\newblock Access via \doi{10.1145/3446776}.
\newblock Republication of \citet{Zhang+2017}.

\end{thebibliography}
\nocite{Trakhtenbrot1984}
\addcontentsline{toc}{section}{References}

\clearpage
\appendix

\part*{Appendices}
\addcontentsline{toc}{part}{Appendices}

The following appendices contain additional results, proofs, and discussion,
as follows.

In \cref{apx:proofs},
    we provide full proofs for the algorithm correctness theorems
    \ref{thm:algo:lossless-compression} and
    \ref{thm:algo:greedy-prank-bound} we
    sketched in the main paper.

In \cref{apx:brank},
    we invert our optimal lossless compression algorithm to offer a
    characterisation of the subset of the parameter space containing
    parameters with a given maximum rank as a union of linear subspaces.

In \cref{apx:prank},
    we document some additional basic properties of the proximate rank.
    First, we prove that the proximate rank varies as a
        right-continuous, non-increasing function of the uniform radius
        $\eps$.
    Second, we show that the proximate rank of two functionally
        equivalent losslessly compressible parameters may vary, but the
        proximate rank of two functionally equivalent incompressible
        parameters cannot.

In \cref{apx:upc},
    we show that \probUPC{} is one of three equivalent perspectives on the
    same abstract decision problem, with each perspective suggesting distinct
    connections to the literature in computational complexity theory.

In \cref{apx:complexity},
    we prove \cref{thm:upc-npc} by reduction from Boolean satisfiability.

In \cref{apx:variants},
    we comment on the complexity of several closely-related problem variants.
    Namely,
        we show that higher-dimensional versions of \probUPC{} are still
        \NP-complete,
        though the problem of covering scalars is in \Poly.
    Moreover, we observe that the reduction in \cref{apx:complexity} suffices
    to show that clique partition on `square penny' graphs is \NP-complete.

In \cref{apx:unbiased},
    we give an alternative proof of the complexity of computing the proximate
    rank, based on exploiting the second of the two opportunities for
        \cref{algo:greedy-prank-bound}
    to be suboptimal listed in \cref{sec:complexity}.
    In particular, we show that the problem of computing the proximate rank
    for an architecture without biases (where the reduction from \probUPC{}
    is no longer available) is still \NP-complete by reduction from the known
    \NP-complete problem \emph{subset sum}.

In \cref{apx:multidim},
    we outline how all of our results can be extended from networks with
    scalar inputs and outputs to networks with multi-dimensional inputs and
    outputs.

\clearpage
\section{Proofs of algorithm correctness theorems}
\label{apx:proofs}

Here we provide proofs for
    \cref{thm:algo:lossless-compression,thm:algo:greedy-prank-bound},
completing the sketches from the main paper.

\thmAlgoLosslessCompression*

\begin{myproof}
(i):
    Following the steps of the algorithm we rearrange the summation defining
    $f_w$ to have the form of $f_{w'}$. For each $x\in\Reals^n$,
    \begin{align*}
        f_w(x)
            &   = d + \sum_{i = 1}^h a_i \tanh(b_i x + c_i)
        \\  &   = d + \sum_{i \notin I} a_i \tanh(c_i)
                    + \sum_{i \in I}    a_i \tanh(b_i x + c_i)
                & \text{(cf.~line 3)}
        \\  &   = \delta + \sum_{i \in I} a_i \tanh(b_i x + c_i)
                & \text{(cf.~line 4)}
        \\  &   = \delta
                + \sum_{j=1}^J \sum_{i\in\Pi_j} a_i \tanh(b_i x + c_i)
                & \text{(cf.~line 6)}
        \\  &   = \delta
                + \sum_{j=1}^J \sum_{i\in\Pi_j}
                    \sign(b_i) \cdot a_i
                    \tanh(\sign(b_i) \cdot b_i x + \sign(b_i) \cdot c_i)
                & \text{($\tanh$ odd)}
        \\  &   = \delta
                + \sum_{j=1}^J
                    \left(\sum_{i\in\Pi_j} \sign(b_i) \cdot a_i \right)
                    \tanh(\beta_j x + \gamma_j)
                & \text{(cf.~lines 6, 9)}
        \\  &   = \delta
                + \sum_{j=1}^J \alpha_j \tanh(\beta_j x + \gamma_j)
                & \text{(cf.~line 8)}
        \\  &   = \delta
                + \sum_{j=1}^r
                    \alpha_{k_j} \tanh(\beta_{k_j} x + \gamma_{k_j})
                & \text{(cf.~line 12)}
        \\  &   = f_{w'}(x).
                & \text{(cf.~line 14)}
    \end{align*}

(ii):
    Each of the reducibility conditions fails to hold for $w' \in \W_r$:
    \ref{item:reducibility:1}~no $\alpha_k$ is zero, due to line~12;
    \ref{item:reducibility:2}~no $\beta_k$ is zero, due chiefly to line~3;
    \ref{item:reducibility:3}, \ref{item:reducibility:4}~all
        $\pm (\beta_k, \gamma_k)$ are distinct, due chiefly to line~6.
\end{myproof}

\thmAlgoGreedyPrankBound*

\begin{myproof}
    Trace the algorithm to construct a parameter $u \in \cnbhd\eps{w}$
    with $\rank(u) = \CALL{Bound}(\eps,w)$.
    Construct $u = (\prange{a^{(u)}}{b^{(u)}}{c^{(u)}},d) \in \W_h$
    as follows.
    \begin{enumerate}
        \item
            For $i \notin I$, $\cnorm{b_i} \leq \eps$, so set
                $b^{(u)}_i = 0$,
            leaving $a^{(u)}_i = a_i$ and $c^{(u)}_i = c_i$.
        \item
            For $i \in \Pi_j$, note that
                $\cdist{\sign(b_i) \cdot (b_i, c_i)}{v_j} \leq \eps$,
            so set $(b_i^{(u)},c_i^{(u)}) = \sign(b_i) \cdot v_j$.
        \item
            For $i \in \Pi_j$, if
                $\cnorm{\alpha_j} \leq \eps\cdot\cardinality{\Pi_j}$,
            then set
                $\displaystyle
                    a_i^{(u)}
                    =
                    a_i - \sign(b_i)\cdot\frac{\alpha_j}{\cardinality{\Pi_j}}
                $,
            else set $a_i^{(u)} = a_i$.
    \end{enumerate}

    By construction, $u \in \cnbhd\eps{w}$.
    To see that $\rank(u) = \CALL{Bound}(\eps,w)$, run \cref{algo:rank} on
    $u$:
        Stage~1 finds the same $I$, since those $b_i^{(u)} = 0$.
        Stage~2 finds the same $\subrange1J\Pi$, since
            $
                \sign(b_i^{(u)}) \cdot (b_i^{(u)}, c_i^{(u)})
                = \sign(b_i) \cdot (b_i^{(u)}, c_i^{(u)})
                = v_j
            $
    ($\sign(b_i^{(u)}) = \sign(b_i)$ since the first component of $v_j$ is
    positive, by line~5 of \cref{algo:greedy-prank-bound}).
    Finally, Stage~3 excludes the same $\alpha_j$, since for $j$ such that
        $\cnorm{\alpha_j} \leq \eps\cdot \cardinality{\Pi_j}$,
    these units from $u$ merge into one unit with outgoing weight
        \begin{equation*}
            \sum_{i\in\Pi_j} \sign(b_i^{(u)}) \cdot a_i^{(u)}
            = \sum_{i\in\Pi_j} \sign(b_i) \cdot \left(
                a_i
                - \sign(b_i) \cdot \frac{\alpha_j}{\cardinality{\Pi_j}}
            \right)
            = \sum_{i\in\Pi_j} \sign(b_i) \cdot a_i - \alpha_j
            = 0
            .
        \qedhere
        \end{equation*}
\end{myproof}

\clearpage
\section{A characterisation of the class of bounded-rank parameters}
\label{apx:brank}

We offer a characterisation of the subset of parameter space with parameters
of a given maximum rank. These are the subsets to which detecting proximity
is proved \NP-complete in \cref{thm:par-npc}.

Let $r, h \in\Natz$ with $r \leq h$. The \emph{bounded rank region} of rank
$r$ is the subset of parameters of rank at most $r$, 
    $\Brank{r} = \Set{w \in \W_h}{\rank(w) \leq r} \subseteq \W_h$.
The key to characterising bounded rank regions is that for each parameter in
    $\Brank{r}$,
at least $h-r$ units would be removed during lossless compression.
Considering the various possible ways in which units can be removed in the
course of \cref{algo:lossless-compression} leads to a characterisation of the
bounded rank region as a union of linear subspaces.

To this end, let $H = \setrange1h$, and define a \emph{compression trace on $h$
units} as a 4-tuple
    $(\bar I, \Pi, \bar K, \sigma)$
where
    $\bar I \subseteq H$ is a subset of units
        (conceptually, those to be removed in Stage~1),
    $\Pi = \subrange1J\Pi$ is a partition of $H \setminus \bar I$
        (the remaining units) into $J$ groups (to be merged in Stage~2),
    $\bar K \subseteq \setrange1J$
        (merged units removed in Stage~3),
and $\sigma \in \set{-1,+1}^h$ is a sign vector
        (unit orientations for purposes of merging).
The length of the compression trace $(\bar I, \Pi, \bar K, \sigma)$ on $h$ units is
    $J - \cardinality{\bar K}$ (representing the number of units remaining).
A compression trace of length $r$ thus captures the notion of a ``way in which
$h-r$ units can be removed in the course of \cref{algo:lossless-compression}''.
\begin{theorem}
    \label{thm:brank-characterisation}
    Let $r \leq h\in\Natz$.
    The bounded rank region $\Brank{r} \subseteq \W_h$ is a union of linear
    subspaces
    \begin{equation}
        \label{eq:brank-characterisation}
        \Brank{r}
        = \bigcup_{(\bar I,\Pi,\bar K,\sigma)\in\RS[h]{r}}
            \left(
                  \bigcap_{i \in \bar I}    S^{(1)}_i
              \cap\bigcap_{j = 1}^{J}       S^{(2)}_{\Pi_j,\sigma}
              \cap\bigcap_{k \in \bar K}    S^{(3)}_{\Pi_k,\sigma}
            \right)
    \end{equation}
    where $\RS[h]{r}$ denotes the set of all compression traces on $h$ units
    with length $r$;
    \begin{align*}
        S^{(1)}_i
            &=  \Set{(\abcrange,d)\in\W_h}{b_i = 0}
        ;
    \\  S^{(2)}_{\Pi,\sigma}
            &=  \Set{(\abcrange,d)\in\W_h}{\forall i,j \in \Pi,\,
                    \sigma_i b_i=\sigma_j b_j \land \sigma_i c_i=\sigma_j c_j
                }
        ;\text{ and}
    \\  S^{(3)}_{\Pi,\sigma}
            &=  \Set{(\abcrange,d)\in\W_h}{\textstyle
                    \sum_{i\in\Pi} \sigma_i a_i = 0
                }.
    \end{align*}
\end{theorem}

\begin{myproof}

($\supseteq$):
    Suppose $w = (\abcrange,d) \in \W_h$ is in the union in
        (\ref{eq:brank-characterisation}),
    and therefore in the intersection for some compression trace 
        $(\bar I, \Pi, \bar K, \sigma)$.
    The constraints imposed on $w$ by membership in this intersection imply
    that the network is compressible:
    \begin{enumerate}
        \item
            For $i\in\bar I$, since $w \in S^{(1)}_i$, $b_i=0$,
            so unit $i$ can be removed.
        \item
            For $j = \range1J$, since $w \in S^{(2)}_{\Pi_j,\sigma}$,
            the units in $\Pi_j$ can be merged together.
        \item
            For $k \in \bar K$, since $w \in S^{(3)}_{\Pi_k,\sigma}$,
            merged unit $k$ has outgoing weight~$0$ and can be removed.
    \end{enumerate}
    It follows that there is a parameter with $J - \cardinality{\bar K}$
    units that is functionally equivalent to $w$.
    Therefore $\rank(w) \leq J - \cardinality{\bar K} = r$ and
    $w \in \Brank{r}$.

($\subseteq$):
    Conversely, suppose $w \in \Brank{r}$.
    Construct a compression trace following $\CALL{Compress}(w)$.
    First, set $\sigma_i = \sign(b_i)$ where $b_i \neq 0$ (if $b_i=0$, set
    $\sigma_i=\pm1$ arbitrarily, this has no effect).
    Then:
    \begin{enumerate}
        \item
            Set $\bar I = \setrange1h \setminus I$ where $I$ is computed on
            line~3. It follows that for $i \in \bar I$, $w \in S^{(1)}_i$.
        \item
            Set $\Pi$ to the partition computed on line~6.
            It follows that for $j=\range1J$, $w \in S^{(2)}_{\Pi_j,\sigma}$.
        \item
            Set $\bar K = \Set{j\in\setrange1J}{\alpha_j=0}$ (cf.~lines~8,12).
            Thus for $k\in\bar K$, $w \in S^{(3)}_{\Pi_k,\sigma}$.
    \end{enumerate}
    By construction $w$ is in
        $
                  \bigcap_{i \in \bar I}    S^{(1)}_i
              \cap\bigcap_{j = 1}^{J}       S^{(2)}_{\Pi_j,\sigma}
              \cap\bigcap_{k \in \bar K}    S^{(3)}_{\Pi_k,\sigma}
        $.
    However, the compression trace $(\bar I, \Pi, \bar K, \sigma)$ has length 
        $\rank(w) \leq r$.
    If $\rank(w) < r$, remove constraints on $r - \rank(w)$ units by some
    combination of the following operations:
        (1)~remove one unit from $\bar I$ (add it as singleton group to $\Pi$),
        (2)~remove one unit from a non-singleton group in $\Pi$ (add it back
            to $\Pi$ as a singleton group),
    and/or
        (3)~remove one merged unit from $\bar K$.
    None of these operations add non-trivial constraints on $w$, so it's still
    the case that $w$ is in the intersection for the modified compression
    trace.
    However, now the length of the compression trace is $r$, so it follows
    that $w$ is in the union as required.
\end{myproof}

\clearpage
\section{Some additional properties of the proximate rank}
\label{apx:prank}

We document some additional basic properties of the proximate rank.

\paragraph{Variation with uniform radius.}

Fix $w \in \W_h$. Then $0 \leq \prank\eps(w) \leq \rank(w)$, depending on
$\eps$.
The following proposition demonstrates some basic properties of this
relationship.

\begin{proposition}
    Let $h \in \Natz$.
    Fix $w \in \W_h$ and consider $\prank\eps(w)$ as a function of $\eps$.
    Then,
    \begin{enumerate}[label=(\roman*)]
        \item
            $\prank\eps(w)$ is antitone in $\eps$:
            if $\eps \geq \eps'$ then
                $\prank\eps(w) \leq \prank{\eps'}(w)$;
            and
        \item
            $\prank\eps(w)$ is right-continuous
            in $\eps$:
            $
                \lim_{\delta \to 0^{+}} \prank{\eps+\delta}(w)
                =
                \prank\eps(w)
            $.
    \end{enumerate}
\end{proposition}

\begin{myproof}
    For (i), put $u \in \cnbhd{\eps'}{w}$ such that
    $\rank(u) = \prank{\eps'}(w)$. Then $u \in \cnbhd\eps{w}$,
    so $\prank{\eps'}(w) = \rank(u) \geq \prank\eps(w)$.
    Then for (ii), since $\prank\eps(w) \leq \rank(w)$ the limit exists by
    the monotone convergence theorem.
    Proceed to bound $\prank\eps(w)$ above and below by
        $r = \lim_{\delta \to 0^{+}} \prank{\eps+\delta}(w)$.
    For the lower bound, for $\delta\in\PosReals$, 
        $\prank{\eps+\delta}(w) \leq \prank\eps(w)$ by (i),
    so
        $r = \lim_{\delta \to 0^{+}} \prank{\eps+\delta}(w) \leq
        \prank\eps(w)$.

    For the upper bound, since the proximate rank is a natural number,
    the limit is achieved for some positive $\delta$.
    That is,
        $\exists\Delta\in\PosReals$ such that $\prank{\eps+\Delta}(w) = r$.
    Then for $k = 1, 2, \ldots$ put
        $u_k \in \cnbhd{\eps+\Delta/k}{w}$
    with $\rank(u_k) = r$.
    Since $\cnbhd{\eps+\Delta}{w}$ is compact the sequence
        $u_1, u_2, \ldots$
    has an accumulation point---call it $u$.
    Now, since $u_1, u_2, \ldots \in \Brank{r}$
        (parameters of rank at most $r$, \cref{apx:brank}),
    a closed set (by \cref{thm:brank-characterisation}),
    the accumulation point $u \in \Brank{r}$.
    Thus, $\rank(u) \leq r$.
    Finally,
        $u \in \bigcap_{k=1}^\infty \cnbhd{\eps+\Delta/k}{w} =
        \cnbhd\eps{w}$,
    so
        $r \geq \rank(u) \geq \prank\eps(w)$.
\end{myproof}

A similar proof implies that $\prank\eps(w)$ achieves its upper bound
$\rank(w)$ for small enough $\eps > 0$.
The lower bound $0$ is also achieved; for example for
    $\eps \geq \cnorm{w} = \cdist{w}{0}$.

\paragraph{Variation with functionally equivalent parameters.}

The rank of $w \in \W_h$ is defined in terms of $f_w$, so functionally
equivalent parameters have the same rank. This is not necessarily the case
for the proximate rank---consider \cref{eg:prank-dependence}.
Similar counterexamples hold if a non-uniform metric is used.
This is a consequence of the fundamental observation that functionally
equivalent parameters may have rather different parametric neighbourhoods.
However, functionally equivalent \emph{incompressible} parameters have the
same proximate rank (\cref{thm:prank-isometries,thm:prank-incompressible}).

\begin{example}
    \label{eg:prank-dependence}
    Let $\eps\in\PosReals$.
    Consider the neural network parameters $w, w' \in \W_2$ with
        $w  = (2\eps, 2\eps, 0, 0, 0, 0, 0)$
    and 
        $w' = (2\eps, 2\eps, 0, 5\eps, 0, 0, 0)$.
    Then for $x\in\Reals$,
        \begin{align*}
            f_w(x) &    = 0 + 2\eps \tanh(2\eps x + 0) + 0 \tanh(    0 x + 0)
            \\     &    = 0 + 2\eps \tanh(2\eps x + 0) + 0 \tanh(5\eps x + 0)
                        = f_{w'}(x),
        \end{align*}
    but
        $
            \prank\eps(w)
             = \rank(\eps, \eps, 0; -\eps, \eps, 0; 0)
             = 0 \neq 1
             = \prank\eps(w')
        $.
\end{example}

\begin{proposition}
    \label{thm:prank-isometries}
    Let $h \in \Natz$.
    Consider a permutation $\pi \in S_h$ and a sign vector $\sigma \in
    \set{-1,+1}^h$.
    Define $T_{\pi,\sigma} : \W_h \to \W_h$ such that
        \begin{equation*}
            T_{\pi,\sigma}(\abcrange,d) = (
                \sigma_1 a_{\pi(1)}, \sigma_1 b_{\pi(1)}, \sigma_1 c_{\pi(1)},
                \ldots,
                \sigma_h a_{\pi(h)}, \sigma_h b_{\pi(h)}, \sigma_h c_{\pi(h)},
                d
            ).
        \end{equation*}
    Then for $w \in \W_h, \eps\in\PosReals$,
        $
            \prank\eps(w) = \prank\eps(T_{\pi,\sigma}(w'))
        $.
\end{proposition}

\begin{myproof}
    $T_{\pi,\sigma} : \W_h \to \W_h$ is an isometry with respect to the
    uniform distance, so
        \begin{equation*}
            \cnbhd\eps{T_{\pi,\sigma}(w)}
                = \Set{T_{\pi,\sigma}(u)}{u \in \cnbhd\eps{w}}.
        \end{equation*}
    Moreover, $T_{\pi,\sigma}$ is a symmetry of the parameter--function map
        \citep{Chen+1993},
    so it preserves the implemented function and therefore the rank of the
    parameters in the neighbourhood.
\end{myproof}

\begin{corollary}
    \label{thm:prank-incompressible}
    Let $h\in\Natz$. 
    Consider two incompressible parameters
        $w, w' \in \W_h$
    and a uniform radius $\eps\in\PosReals$.
    If $f_w = f_{w'}$ then $\prank\eps(w) = \prank\eps(w')$.
\end{corollary}

\begin{myproof}
    Since $w, w'$ are functionally equivalent and incompressible,
    they are related by a permutation transformation and a negation
    transformation
        \citep{Sussmann1992}.
\end{myproof}

\clearpage
\section{Three perspectives on uniform point cover, and related problems}
\label{apx:upc}

\begin{figure}[h!]
    \centering
    \includegraphics[width=\textwidth]{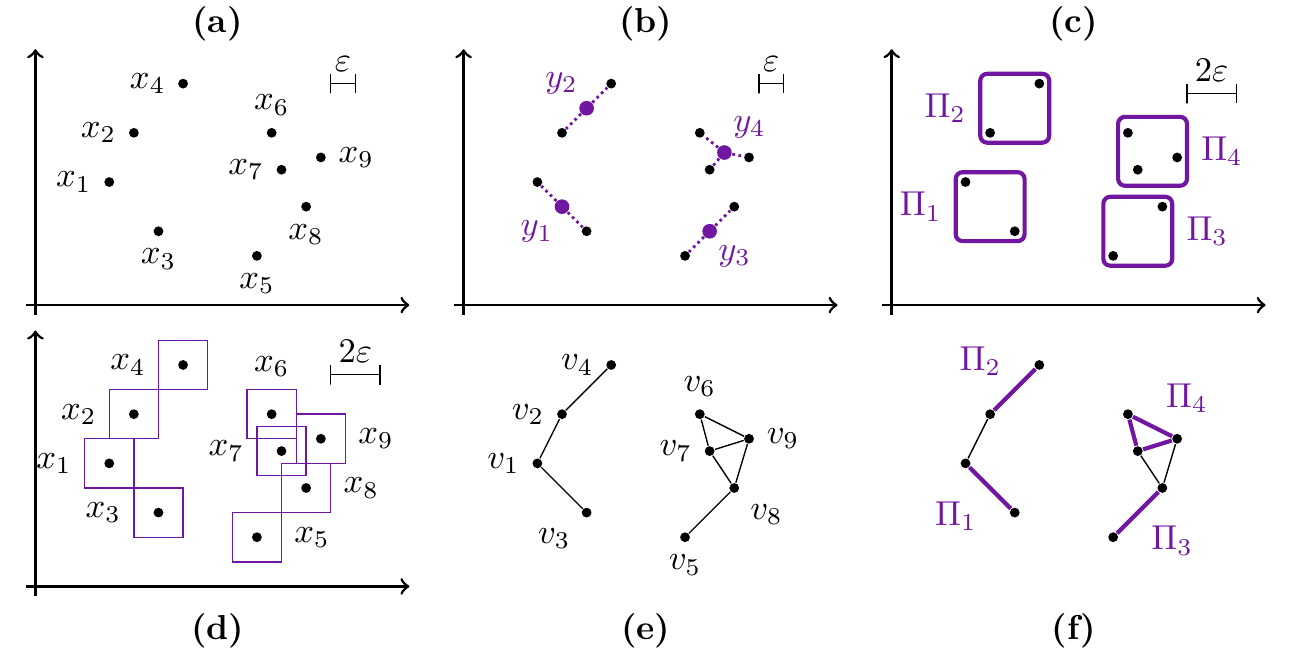}
    \caption{%
        \label{fig:three-perspectives}
        Example instances for \probUPC{}, \pref{prob:upp}{\UPP}, and
        \pref{prob:usgcp}{\usgCP}.
        \textbf{(a)}~Nine (source) points $\subrange19x$.
        \textbf{(b)}~A $(4, \eps)$-cover $\subrange14y$.
        \textbf{(c)}~A $(4, 2\eps)$-partition $\subrange14\Pi$
                ($= \set{1,3}, \set{2,4}, \set{5,8}, \set{6,7,9}$).
        \textbf{(d)}~The nine points $\subrange19x$, along with $2\eps$-width
            squares.
        \textbf{(e)}~The corresponding unit square graph with vertices
            $\subrange19v$.
        %
        \textbf{(f)}~A partition of the unit square graph into four cliques
            $\subrange14\Pi$.
        \textbf{Note:}~$\eps$ represents a radius in \probUPC{}, but a
        diameter in Problems~\hyperref[prob:upp]{\UPP} and
            \hyperref[prob:usgcp]{\usgCP}.
        In these examples, we use $\eps$ for the radius and $2\eps$ for
        the diameter.
    }
\end{figure}

We show that \probUPC{} is one of three equivalent perspectives on the same
abstract decision problem. Each of the three perspectives suggests distinct
connections to existing problems.
\Cref{fig:three-perspectives} shows example instances of the three problems.

\paragraph{Perspective 1: Uniform point cover.}

Given some points in the plane, is there a small number of new ``covering''
points so that each (original) point is near a covering point?
This is the perspective introduced in the main paper as \probUPC{}. It
emphasises the existence of covering points, which are useful in the
reduction to \probPAR{} for the proof of \cref{thm:par-npc}.

Uniform point cover is reminiscent of well-known hard clustering problems
such as \emph{Planar $k$-means} \citep{Mahajan+2012}.
The $k$-means problem concerns finding $k$ centroids with a low
\emph{sum} of (squared Euclidean) distances between source points
and their nearest centroids.
In contrast, uniform point cover concerns finding covering points with a
low \emph{maximum} (uniform) distance between source points and their
nearest covering points.

\probUPC{} is more similar to the \NP-complete problem of
    \emph{absolute vertex $p$-centre}
    \citep{Hakimi1964,Hakimi1965,Kariv+Hakimi1979}.
This problem concerns finding a set of points on a graph
    (i.e., vertices or points along edges)
with a low maximum (shortest path) distance between vertices and their
nearest points in the set.
\probUPC{} is a geometric $p$-centre problem using uniform distances.

A geometric $p$-centre problem using Euclidean distances was shown to be
\NP-complete by
    \citet[\S4.3.2; see also \citealp{Megiddo+Supowit1984}]{Supowit1981}.
\citet{Megiddo+Supowit1984} also showed the \NP-hardness of an optimisation
variant using $L_1$ distance (rectilinear or Manhattan
distance).\footnotemark{}
We prove that \probUPC{} is \NP-complete using somewhat similar reductions,
but many simplifications afforded by starting from a more restricted variant
of Boolean satisfiability.

\footnotetext{%
    The $L_1$-distance variant is equivalent to \probUPC{} by a $45^\circ$
    rotation of the plane.
}

\paragraph{Perspective 2: Uniform point partition.}

Given some points in the plane, can the points be partitioned into a small
number of groups with small uniform diameter?
This perspective, formalised as \pref{prob:upp}{\UPP}, emphasises the
grouping of points, rather than the specific choice of covering points.

Consider $h$ points $\subrange1hx \in \Reals^2$.
A \emph{$(r,\eps)$-partition} of the $h$ points is a partition of
$\setrange1h$ into $r$ subsets $\subrange1r\Pi$ such that the uniform
distance between points in any subset is at most $\eps$:
$
    \forall k \in \set{1, \ldots, r}\!,
        \forall i, j \in \Pi_k,
            \cdist{x_i}{x_j} \leq \eps
$.

\begin{problem}[\UPP]
    \label{prob:upp}
    \emph{Uniform point partition} (\UPP) is a decision problem.
    Each instance comprises a collection of $h \in \Natz$ points
        $\subrange1hx \in \Reals^2$,
    a uniform diameter $\eps\in\PosReals$,
    and a number of groups $r \in \Natz$.
    Affirmative instances are those for which there exists an
    $(r,\eps)$-partition of $\subrange1hx$.
\end{problem}

\paragraph{Perspective 3: Clique partition for unit square graphs.}

Given a special kind of graph called a \emph{unit square graph,} can the
vertices be partitioned into a small number of cliques?
The third perspective strays from the simple neural network context,
but reveals further related work.

Consider $h$ points in the plane, $\subrange1hx \in \Reals^2$, and a diameter
$\eps \in \PosReals$.
Thus define an undirected graph $(V,E)$ with vertices $V = \setrange1h$ and edges
    $E = \Set{\set{i, j}}{i \neq j, \cdist{x_i}{x_j} \leq \eps}$.
A \emph{unit square graph}\footnotemark{} is any graph that can be
constructed in this way.
Unit square graphs are a uniform-distance variant of unit disk graphs
    \citep[based on Euclidean distance; cf.][]{Clark+1990}.

\footnotetext{%
    ``Unit square graph'' comes from an equivalent definition of these graphs
    as intersection graphs of unit squares.
    To see the equivalence, scale the collection of squares by $\eps$ and
    then consider their centres.
    The same idea relates the \emph{proximity} and \emph{intersection} models
    for unit disk graphs \citep{Clark+1990}.
}
    
Consider an undirected graph $(V,E)$. A \emph{clique partition} of size $r$ is
a partition of the vertices $V$ into $r$ subsets $\subrange1r\Pi$ such that
each subset is a clique:
    $
        \forall k \in \set{1, \ldots, r}\!,
            \forall v_i \neq v_j \in \Pi_k,
                \set{v_i, v_j} \in E
    $.

\begin{problem}[\usgCP]
    \label{prob:usgcp}
    \emph{Clique partition for unit square graphs} (\usgCP) is a decision
    problem.
    Each instance comprises a unit square graph $(V,E)$ and a number of
    cliques $r\in\Natz$.
    Affirmative instances are those for which there exists a clique
    partition of size $r$.
\end{problem}
   
The clique partition problem is \NP-complete in general graphs
\citep{Karp1972}.
\citet{Cerioli+2004,Cerioli+2011} showed that it remains \NP-complete when
restricted to unit disk graphs, using a reduction from a variant of Boolean
satisfiability that is somewhat similar to our reduction.

\newpage

\paragraph{Equivalence of the three perspectives.}

Problems
    \hyperref[prob:upc]{\UPC},
    \hyperref[prob:upp]{\UPP},
and \hyperref[prob:usgcp]{\usgCP}
are equivalent in the sense that there is an immediate reduction between any
pair of them.

\begin{theorem}[Equivalence of Problems \UPC, \UPP, and \usgCP]
    \label{thm:three-perspectives}
    Let $h,r\in\Natz$, $\eps\in\PosReals$, and
        $\subrange1hx \in \Reals^2$.
    The following conditions are equivalent:
    \begin{enumerate}[label=(\roman*)]
        \item
            there exists an $(r, \frac12\eps)$-cover of $\subrange1hx$;
        \item
            there exists an $(r, \eps)$-partition of $\subrange1hx$;
    and \item
        the unit square graph on $\subrange1hx$ (diameter $\eps$)
            has a clique partition of size $r$.
    \end{enumerate}
\end{theorem}

\newcommand\xmaxp{\max_{i \in \Pi_j} x_{i,p}}
\newcommand\xminp{\min_{i \in \Pi_j} x_{i,p}}

\begin{proof}
(ii $\limp$ i):
    Let $\subrange1r\Pi$ be an $(r, \eps)$-partition of the points
    $\subrange1hx$.
    For each $\Pi_j$, define $y_j \in \Reals^2$ as the centroid of the
    bounding rectangle of the set of points $\Set{x_i}{i\in\Pi_j}$,
    that is,
        $
            y_j = \frac12\left(
                \max_{i \in \Pi_j} x_{i,1} + \min_{i \in \Pi_j} x_{i,1}
                ,\ 
                \max_{i \in \Pi_j} x_{i,2} + \min_{i \in \Pi_j} x_{i,2}
            \right)
        $.
    For each $i\in\Pi_j$ and $p\in\set{1,2}$,
    let $\alpha = \argmax_{a\in\Pi_j} x_{a,p}$
    and $\beta  = \argmin_{b\in\Pi_j} x_{b,p}$.
    Then,
    \begin{align*}
        \abs(2x_{i,p}-2y_{j,p})
            &   = \abs\left( 2x_{i,p} - \left(\xmaxp+\xminp\right) \right)
        \\  &   \leq \abs\left( x_{i,p} - \xmaxp \right)
                   + \abs\left( x_{i,p} - \xminp \right)
            &   \text{(triangle inequality)}
        \\  &   = \xmaxp - \xminp
            &   \hspace{-21.24pt}\text{($\xmaxp \leq x_{i,p} \leq \xminp$)}
        \\  &   = x_{\alpha,p}-x_{\beta,p}
                \leq \cdist{x_\alpha}{x_\beta}
                \leq \eps.
    \end{align*}
    Thus $\cdist{x_i}{y_p} \leq \frac12\eps$, and $\subrange1k{y}$ is an
    $(r,\frac12\eps)$-cover of $\subrange1hx$.

(i $\limp$ iii):
    Let $\subrange1ry \in \Reals^2$ be an $(r, \frac12\eps)$-cover of
        $\subrange1hx$.
    Partition $\set{1, \ldots, h}$ into $\subrange1r\Pi$ by grouping
    points according to the nearest covering point (break ties arbitrarily).
    Then for $i,j\in\Pi_k$, $\set{i, j} \in E$ of the unit square graph,
    since
        $
            \cdist{x_i}{x_j}
            \leq \cdist{x_i}{y_k} + \cdist{y_k}{x_j}
            \leq \frac12\eps+\frac12\eps
            = \eps
        $.
    Thus $\subrange1r\Pi$ is a clique partition.

(iii $\limp$ ii):
    Let $\subrange1r\Pi$ be a clique partition.
    Then for $i, j \in \Pi_k$,
        $\set{i,j}\in E$, and so
        $\cdist{x_i}{x_j} \leq \eps$.
    Thus $\subrange1r\Pi$ is an $(r, \eps)$-partition.
\end{proof}

\clearpage
\section{\NP-completeness of uniform point cover}
\label{apx:complexity}

In this section, we prove \cref{thm:upc-npc}.
By \cref{thm:three-perspectives}, it suffices to prove that \probUPP{} is
\NP-complete. This simplifies the proof by abstracting away the construction
of specific covering vectors given each group of points.

The main part of the proof is a reduction from a restricted variant of
Boolean satisfiability we call \probxSAT,
wherein formulas must have
    (1)~two or three literals per clause,
    (2)~one negative occurrence and one or two positive occurrences
        per literal, and
    (3)~a planar bipartite clause--variable incidence graph.
These restrictions further streamline the proof by reducing number of cases
to be considered in the reduction.
\Cref{apx:complexity:xsat} defines \probxSAT{} and proves it is \NP-complete.

\Cref{apx:complexity:upp} gives the reduction from \probxSAT{} to \probUPP{}
and completes the proof.
Even with the above simplifications, the reduction
is substantial and occupies pages
    \pageref{thm:xsat-reducto-upp}--\pageref{thm:xsat-reducto-upp:end}.

\subsection{Restricted Boolean satisfiability}
\label{apx:complexity:xsat}

Boolean satisfiability is a canonical \NP-complete decision problem
    \citep[see also \citealp{Garey+Johnson1979}]{Cook1971,Levin1973}.
We formalise this problem as follows.
Given
    $n$ \emph{variables,} $\subrange1nv$,
a \emph{Boolean formula (in conjunctive normal form)} is a conjunction of
    $m$ \emph{clauses,} $c_1 \land \cdots \land c_m$,
where each clause is a finite disjunction ($\lor$) of \emph{literals,}
and each literal is either a variable $v_i$ or its negation $\bar v_i$
(called, respectively, a \emph{positive occurrence} or \emph{negative
occurrence} of the variable $v_i$).
A \emph{truth assignment} maps each variable to the value ``true'' or
``false.'' 
The formula is \emph{satisfiable} if there exists a truth assignment
such that the formula evaluates to ``true'', that is, each clause contains at
least one positive occurrence of a ``true'' variable or negative occurrence of
``false'' variable.

\begin{problem}[\SAT]
    \label{prob:sat}
    \emph{Boolean satisfiability} (\SAT) is a decision problem.
    The instances are all Boolean formulas in conjunctive normal form.
    Affirmative instances are all satisfiable formulas.
\end{problem}

We introduce \probxSAT, a variant of \pref{prob:sat}{\SAT}.
Let $\phi$ be a Boolean formulas with variables $\subrange1nv$ and
clauses $c_1 \land \cdots \land c_m$.
Call $\phi$ a \emph{restricted Boolean formula} if it meets the following
three additional conditions.
\begin{enumerate}
    \item
        Each variable $v_i$ occurs as a literal in either two clauses or
        in three clauses. Exactly one of these occurrences is a negative
        occurrence (the other one or two are positive occurrences).
    \item
        Each clause $c_j$ contains either two literals or three literals
        (these may be any combination of positive and negative).
    \item
        The \emph{bipartite variable--clause incidence graph} of $\phi$---%
        an undirected graph
            $(V,E)$
        with vertices $V = \set{v_1, \ldots, v_n, c_1, \ldots, c_m}$
        and edges
            $E = \{\, {\set{v_i, c_j}} \,:\,$
                variable $v_i$ occurs as a literal in clause $c_j$
            $\,\}$%
        ---is a planar graph.
\end{enumerate}

\begin{problem}[\xSAT]
    \label{prob:xsat}
    \emph{Restricted Boolean satisfiability} (\xSAT) is a decision problem.
    The instances are all \emph{restricted} Boolean formulas.
    Affirmative instances are all satisfiable formulas.
\end{problem}

\begin{figure}[h!]
    \centering
    \includegraphics[width=\textwidth]{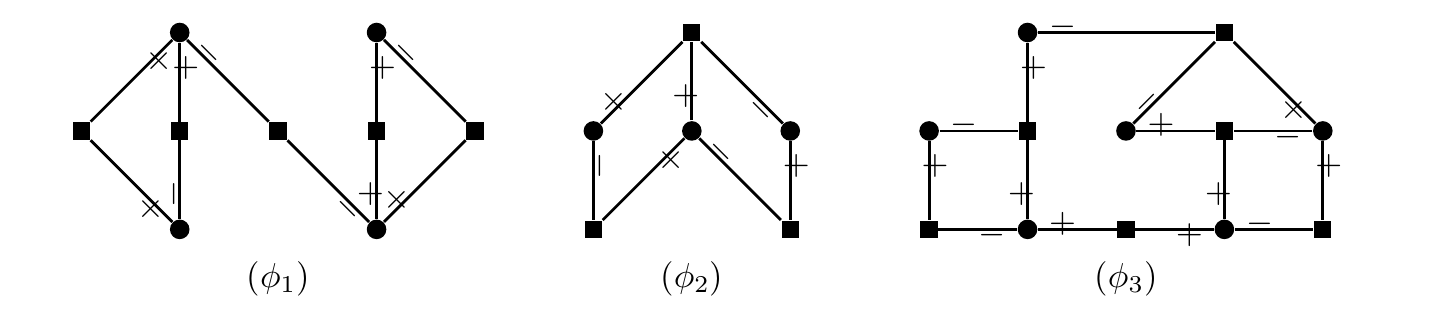}
    \caption{\label{fig:xsat}%
        Three example restricted Boolean formulas in conjunctive normal form
        are as follows:
        $
            \phi_1 =   (v_1 \lor v_2)
                \land   (\bar v_1 \lor v_2)
                \land   (v_3 \lor v_4)
                \land   (\bar v_3 \lor v_4)
                \land   (\bar v_2 \lor \bar v_4)
        $;
        $
            \phi_2 =   (\bar v_1 \lor v_2)
                \land   (v_1 \lor v_2 \lor \bar v_3)
                \land   (\bar v_2 \lor v_3)
        $;
        and 
        $
            \phi_3 =   (\bar v_1 \lor \bar v_3 \lor v_4)
                \land   (v_1 \lor \bar v_2 \lor v_5)
                \land   (v_3 \lor \bar v_4 \lor v_6)
                \land   (v_2 \lor \bar v_5)
                \land   (v_5 \lor v_6)
                \land   (v_4 \lor \bar v_6)
        $.
        The bipartite variable--clause incidence graphs of the three formulas
        are depicted above (circles indicate variable vertices, squares
        indicate clause vertices, positive and negative occurrences (edges)
        are marked accordingly.
        Formula $\phi_1$ is unsatisfiable whereas formulas $\phi_2$ and
        $\phi_3$ are each satisfiable.
    }
\end{figure}

\Cref{fig:xsat} illustrates some restricted Boolean formulas.

\clearpage
\begin{theorem}
    \probxSAT{} is \NP-complete.
\end{theorem}

\begin{myproof}
    $\xSAT \in \NP$ since $\SAT \in \NP$.
    To show $\SAT \reducto \xSAT{}$ much of the work is already done:
    \begin{enumerate}
        \item
            \citet{Cook1971} reduced \SAT{} to \tSAT{}, a variant with
            at most three literals per clause.
        \item
            \citet{Lichtenstein1982} extended this reduction to \ptSAT{}, a
            variant with at most three literals per clause and a planar
            bipartite clause--variable incidence graph
            (in fact the planarity condition studied by Lichtenstein is
            even stronger).
        \item
            \citet{Cerioli+2004,Cerioli+2011} extended the reduction to
            \pttSAT{}---a variant of \ptSAT{} with at most three occurrences
            per variable.
            \citet{Cerioli+2004,Cerioli+2011} used similar techniques to
            \citet{Tovey1984}, \citet{Jansen+Muller1995}, and
            \citet{Berman+2003}, who studied variants of \SAT{} with bounded
            occurrences per variable, but no planarity restriction.
    \end{enumerate}
    It remains to efficiently construct from an instance $\phi$ of \pttSAT{}
    an equisatisfiable formula $\phi'$ having \emph{additionally}
        (a)~at least two occurrences per variable,
        (b)~at least two variables per clause,
    and (c)~exactly one negative occurrence per variable.
    This can be achieved by removing variables, literals, and clauses and
    negating occurrences in $\phi$ (noting that such operations
    do not affect the conditions on $\phi$) as follows.
    First, establish (a) and (b)\footnotemark{} by exhaustively applying
    the following (polynomial-time) operations.
    \footnotetext{\label{fn:empty-clauses}%
        If a clause contains no literals, whether initially or due to
        the removal of a literal through operation (ii), then the formula is
        unsatisfiable.
        Return any unsatisfiable instance of \xSAT{}, such as $\phi_1$ of
        \cref{fig:xsat}.
    }
    \begin{enumerate}[label=(\roman*)]
        \item
            If a variable always occurs with one sign (including never or
            once), remove the variable and all incident clauses.
            The resulting (sub)formula is equisatisfiable:
            extend a satisfying assignment by satisfying the removed clauses
            with the removed variable.
        \item
            If a clause contains a single literal, this variable is
            determined in a satisfying assignment.
            Remove the variable and clause along with any other clauses
            in which the variable occurs with that sign.
            For other occurrences, retain the clause but remove the
            literal.%
                \textsuperscript{\ref{fn:empty-clauses}}
            If the resulting formula is unsatisfiable, then the additional
            variable won't help, and if the resulting formula
            is satisfiable, then so is the original with the appropriate
            setting of the variable to satisfy the singleton clause.
    \end{enumerate}
    Only a polynomial number of operations are possible as each removes
    one variable.
    Moreover, thanks to (i), each variable with two occurrences now has
    one negative occurrence (as required).
    For variables with three occurrences, one or two are negative.
    Establish (c) by negating all three occurrences for those that have two
    negative occurrences
        (so that the two become positive and the one becomes negative
        as required).
    The result is equisatisfiable because satisfying assignments can be
    translated by negating the truth value assigned to this variable.
    Carrying out this negation operation for the necessary variables takes
    polynomial time and completes the reduction.
\end{myproof}

\subsection{Complexity of uniform point partition}
\label{apx:complexity:upp}

\Cref{thm:upc-npc} is a corollary of \cref{thm:three-perspectives} and the
following two results.

\begin{theorem}
    \label{thm:upp-np}
    $\UPP \in \NP$.
\end{theorem}

\begin{myproof}
    An $(r,\eps)$-partition of the points acts as a certificate.
    Such a partition can be verified in polynomial time by computing the
    pairwise uniform distances within each group.
\end{myproof}

\begin{theorem}
    \label{thm:xsat-reducto-upp}
    $\xSAT \reducto \UPP$.
\end{theorem}

\begin{myproof}
This proof is more substantial.
We describe the reduction algorithm in detail, and then formally prove its
efficiency and correctness, over the remainder of this section
(pages \pageref{thm:xsat-reducto-upp}--\pageref{thm:xsat-reducto-upp:end}).

\paragraph{Reduction overview.}

Given an \xSAT{} instance, the idea is to build a \UPP{} instance with
a collection of points mirroring the structure of the bipartite
variable--clause incidence graph of the restricted Boolean formula.
To each variable vertex, clause vertex, and edge corresponds a collection
of points. The points for each variable vertex can be partitioned in one
of two configurations, based on the value of the variable in a truth
assignment. Each determines the available groupings of the incident
edge's points so as to propagate these assignments to the clauses.
The maximum number of groups is set so that there are enough to include
the points of each clause vertex if and only if some variable satisfies
that clause in the assignment.
\Cref{fig:overview} shows one example of this construction.

\begin{figure}[!h]
    \centering
    \includegraphics[width=\textwidth]{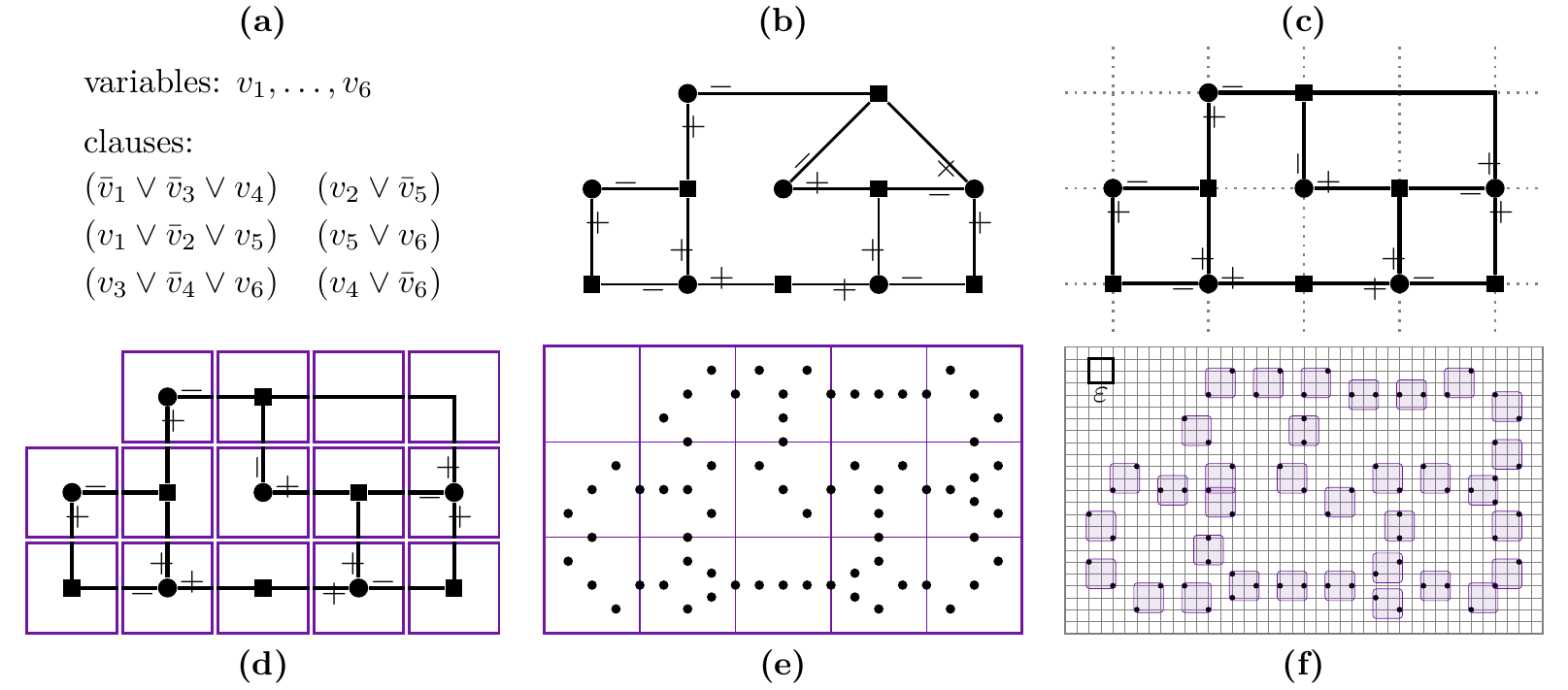}
    \caption{\label{fig:overview}%
        Example of reduction from restricted Boolean satisfiability to
        \probUPP.
        \textbf{(a)}~A satisfiable restricted Boolean formula.
        \textbf{(b)}~The formula's planar bipartite variable--clause
        incidence graph (circles: variables, squares: clauses,
        edges: $\pm$ literals).
        \textbf{(c)}~The graph embedded onto an integer grid.
        \textbf{(d)}~The embedding divided into unit tiles of various types.
        \textbf{(e)}~The $h = 68$ source points aggregated from each of the
        tiles.
        \textbf{(f)}~Existence of a $(34,1/4)$-partition of the source
        points.
    }
\end{figure}

\paragraph{Reduction step 1: Lay out the graph on a grid.}

Due to the restrictions on the \xSAT{} instance, the bipartite
variable--clause incidence graph is planar with maximum degree three.
Therefore there exists a graph layout where
    (1)~the vertices are positioned at integer coordinates, and
    (2)~the edges comprise horizontal and vertical segments between adjacent
        pairs of integer coordinates
    \citep[\S{}IV]{Valiant1981}.
Moreover, such a \emph{(planar, rectilinear, integer) grid layout}
can be constructed in polynomial time
    \citetext{see, e.g., \citealp{Valiant1981,Liu+1998};
        note there is no requirement to produce an ``optimal'' layout---just
        a polynomial-time computable layout%
    }.
\Cref{fig:grid-layout} shows three examples.

\paragraph{Reduction step 2: Divide the layout into tiles.}

The grid layout serves as a blueprint for a \UPP{} instance: it governs
how the points corresponding to each variable vertex, clause vertex, and
edge are arranged in the plane.
The idea is to conceptually divide the plane into unit square
    \emph{tiles,} 
with one tile for each coordinate of the integer grid occupied by a
vertex or edge in the grid layout.
The tile divisions for the running examples are shown in
\cref{fig:tile-division}.

\begin{figure}[h!]
    \centering
    \includegraphics[width=\textwidth]{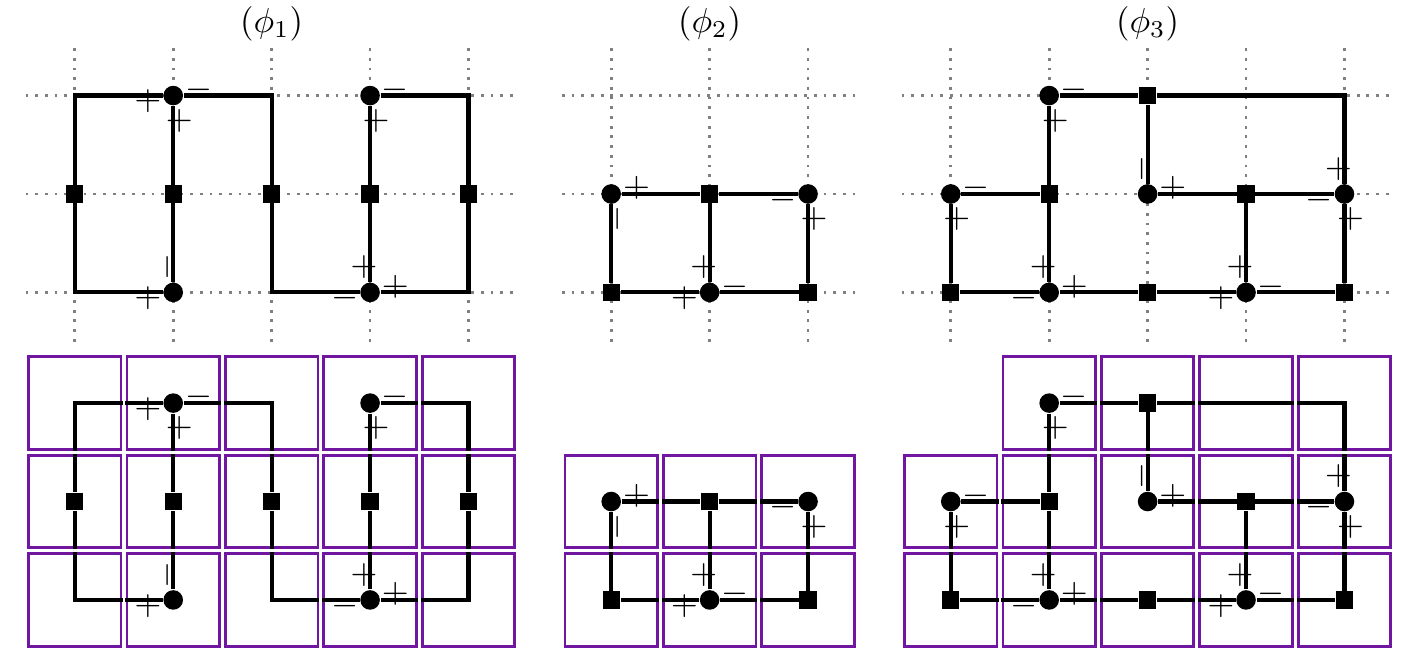}
    \caption{%
        \label{fig:grid-layout}%
        \label{fig:tile-division}%
        \textbf{Top:}
        Example planar, rectilinear, integer grid layouts of the
        bipartite variable--clause incidence graphs from \cref{fig:xsat}.
        Note: these layouts are computed by hand---those produced by standard
        algorithms may be larger.
        \textbf{Bottom:} Division of the same grid layouts into tiles.
    }
\end{figure}

\newpage

Due to the restrictions on \xSAT{} instances, any tile division uses just
forty distinct tile \emph{types} (just nine up to rotation and reflection).
There are straight edge segments and corner edge segments, plus clause and
variable vertices with two or three edges in any direction, and for variable
vertices, exactly one direction corresponds to a negative occurrence.
\Cref{fig:tile-types} enumerates these types.

\begin{figure}[h!]
    \centering
    \includegraphics[width=0.96\textwidth]{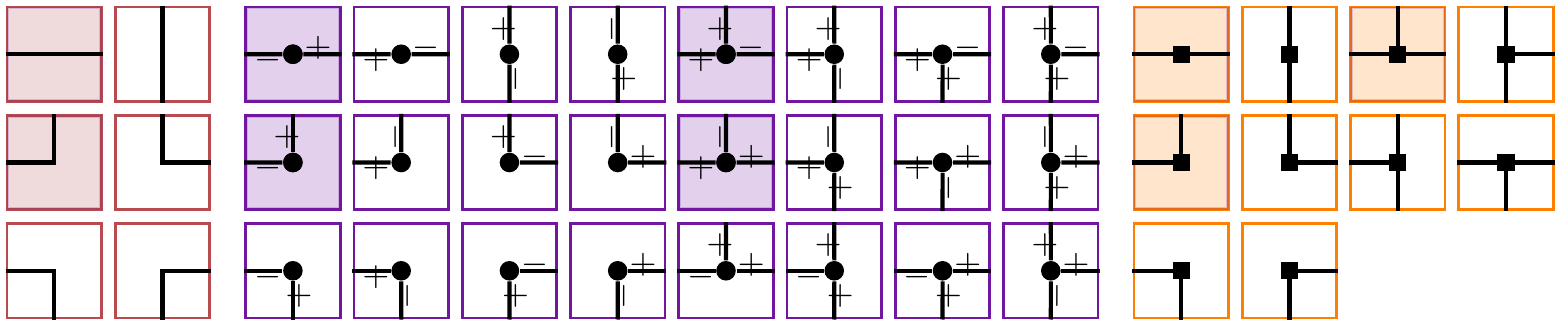}
    \caption{%
        \label{fig:tile-types}%
        Just forty tile types suffice to construct any tile division of a
        grid layout. Up to rotation and reflection, just nine distinct
        types (for example, those highlighted) suffice.
    }
\end{figure}

\paragraph{Reduction step 3: Populate the instance with points.}

The points of the \UPP{} instance are of two kinds (described in more
detail below):
    (1)~\emph{boundary points} between neighbouring pairs of tiles;
and (2)~\emph{interior points} within each tile in a specific arrangement
        depending on the tile type.
The boundary points can be grouped with interior points of one or the other
neighbouring tile. In this way, boundary points couple the choice of how
to partition the interior points of neighbouring tiles, creating the global
constraint that corresponds to satisfiability.

\paragraph{Reduction step 3a: Boundary points between neighbouring tiles.}

There is one boundary point at the midpoint of the boundary between each
pair of neighbouring tiles. A pair of \emph{neighbouring tiles} is one for
which there is an edge crossing the boundary.
It is not sufficient for the tiles to be adjacent.
\Cref{fig:boundary-points} clarifies this distinction using the running
examples.

\begin{figure}[!ht]
    \centering
    \includegraphics{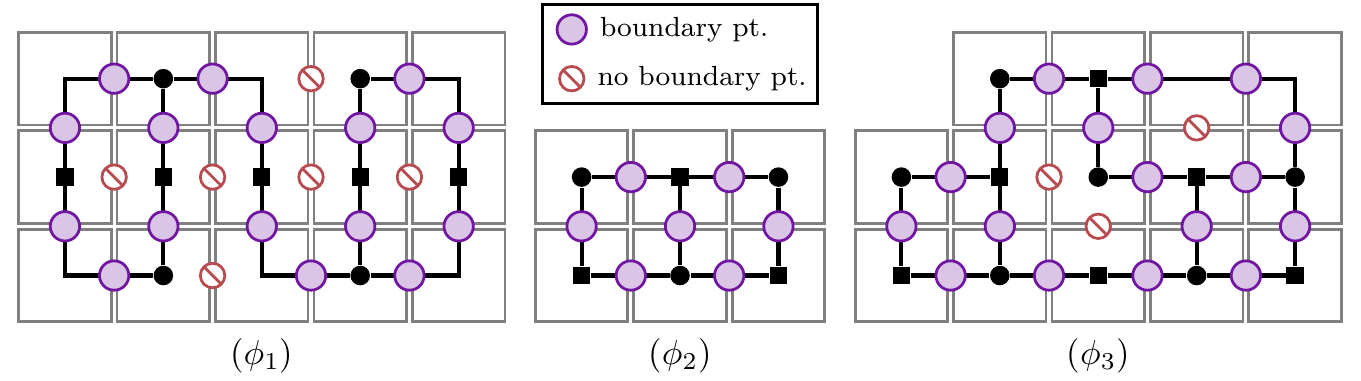}
    \caption{%
        \label{fig:boundary-points}%
        Example of the placement of boundary points between neighbouring
        tiles.
        Boundary points are not placed between adjacent tiles if no edge
        crosses this tile boundary.
    }
\end{figure}

\paragraph{Reduction step 3b: Interior points for variable tiles.}

\Cref{tab:var-tiles} shows arrangements of interior points for each type
of variable tile (up to rotation and reflection).
Due to the restrictions on the \xSAT{} instance, each variable tile
has one \emph{negative boundary point} and one or two \emph{positive boundary
point(s)} (corresponding to the variable occurrences).
With a given number of groups, the choice of which boundary point(s) to
include corresponds to the value of the variable in a truth assignment.
\begin{lemma}
    \label{tilemma:v}
    Consider an interior and boundary point arrangement from
        \cref{tab:var-tiles} (first 4~rows),
    or a rectilinear rotation or reflection of such an arrangement.
    Let $r \in \set{2,3}$ be the allocated number of groups,
    and let $\eps \in \PosReals$ be the scale.
    \begin{enumerate}[label=(\roman*)]
        \item
            There is no $(k,\eps)$-partition of the interior points if
            $k < r$.
        \item
            For any $(r, \eps)$-partition of the interior points,
            the negative boundary point is within uniform distance $\eps$
            of all points in some group,
            if and only if
            (neither of) the positive boundary point(s) are within uniform
            distance $\eps$ of all points in any group.
    \end{enumerate}
\end{lemma}

\begin{myproof}
    It suffices to consider the arrangements in \cref{tab:var-tiles}
    (first 4~rows) because the uniform distance is invariant to rectilinear
    rotation and reflection.
    The claims are then verified by exhaustive consideration of all
    possible partitions of the interior points into at most $r$ groups.
\end{myproof}
   
The partitions indicated in the table are used while constructing a
partition of the whole instance given a satisfying assignment.
Conversely, no other partitions of the interior points using $r$ groups are
possible, except in the fourth row, where other partitions are possible,
but, as suffices for the reduction, there are no partitions including both
positive and negative boundary points.

\begin{table}[h!]
    \centering
    \includegraphics[width=\textwidth]{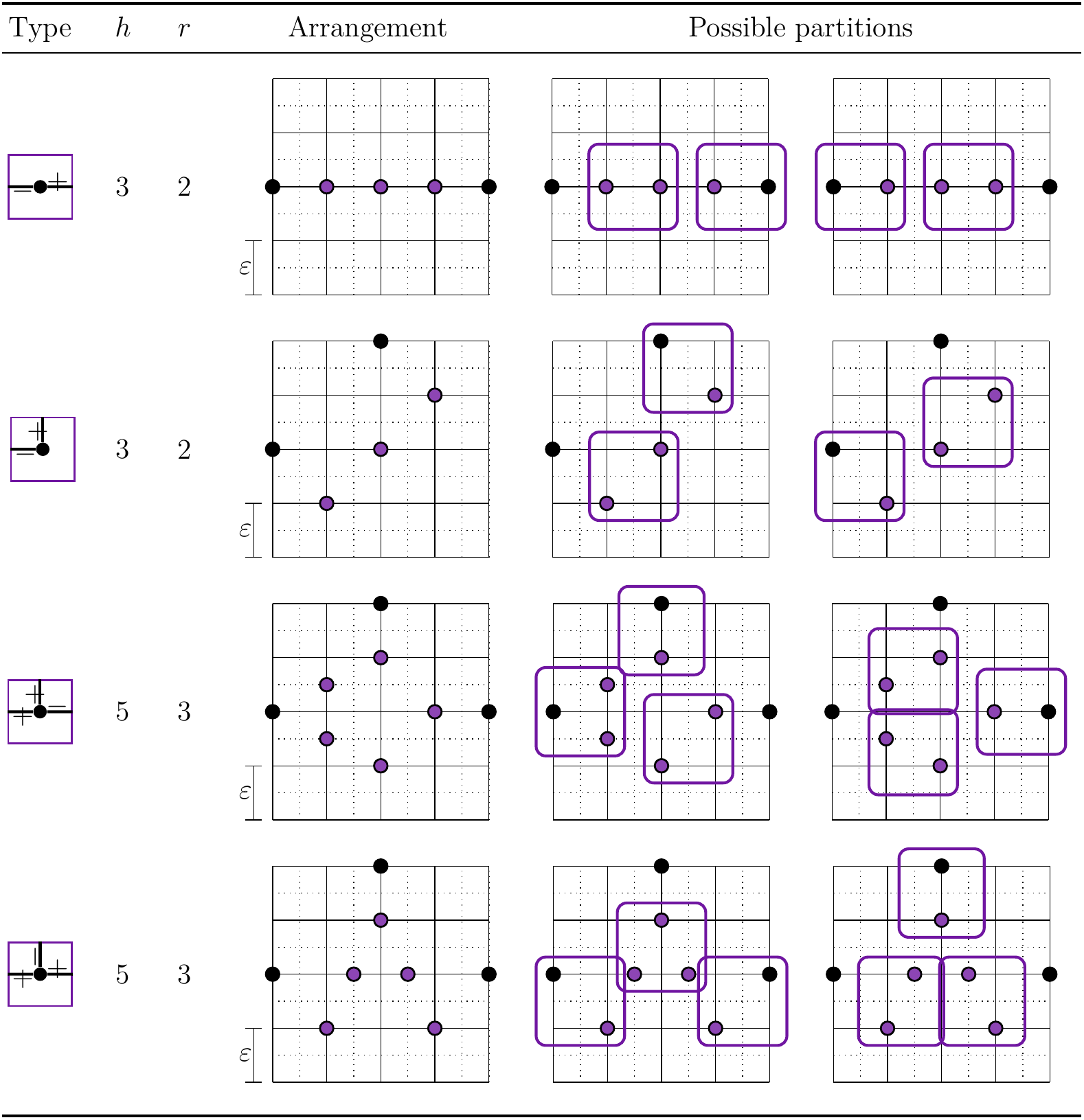}
    \caption{\label{tab:var-tiles}%
        Arrangement of interior points for variable tiles.
        The number $h$ represents the number of interior points (coloured),
        with nearby boundary points also shown (black).
        The number $r$ represents the number of groups allocated to the tile
        during the reduction.
    }
\end{table}

\newpage

\paragraph{Reduction step 3c: Interior points for edge tiles.}

\Cref{tab:edge-tiles} shows arrangements of interior points for each type of
edge tile (up to rotation and reflection).
Once the partition of a variable tile includes either the positive boundary
point(s) or the negative boundary point, the role of an edge tile is to
propagate this choice to the incident clause.
These simple point arrangements ensure that the opposite boundary point can
be included in a partition of the interior points if and only if the prior
boundary point is not (that is, if and only if it was included by the
partition of the interior points of the variable tile or, inductively, the
previous edge tile).

\begin{lemma}
    \label{tilemma:e}
    Consider an interior and boundary point arrangement from
        \cref{tab:edge-tiles} (last 2~rows),
    or a rectilinear rotation or reflection of such an arrangement.
    Let $\eps \in \PosReals$ be the scale.
    \begin{enumerate}[label=(\roman*)]
        \item
            There is no $(k,\eps)$-partition of the interior points if
            $k < 2$.
        \item
            For any $(2, \eps)$-partition of the interior points,
            either boundary point is within uniform distance $\eps$ of all
            points in some group,
            if and only if
            the other boundary point is not within uniform distance $\eps$ of
            all points in any group.
    \end{enumerate}
\end{lemma}

\begin{myproof}
    Special case of \cref{tilemma:v}.
\end{myproof}

The partitions indicated in \cref{tab:edge-tiles} are the only possible
$(2,\eps)$-partitions of the interior points.

\begin{table}[h!]
    \centering
    \includegraphics[width=\textwidth]{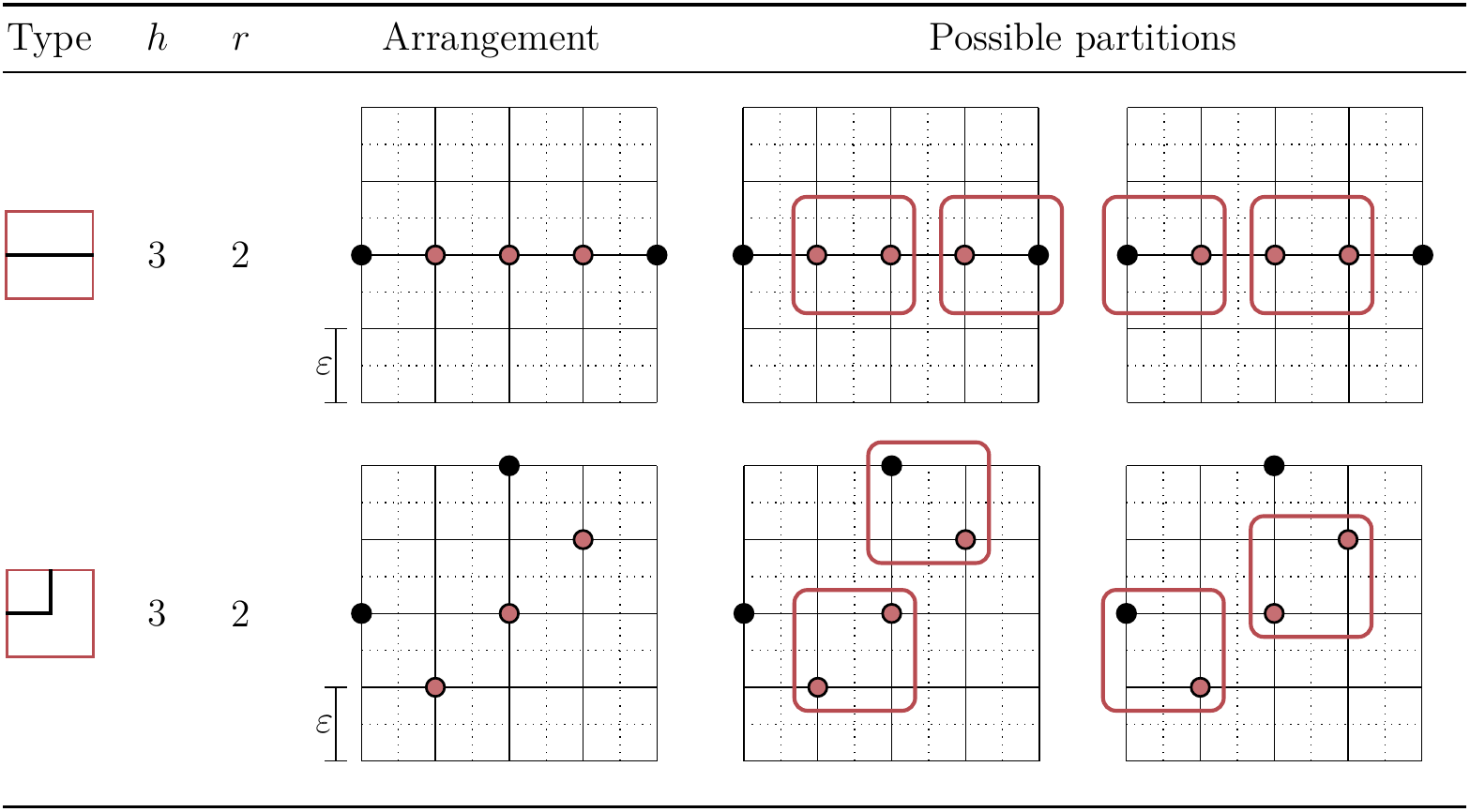}
    \caption{\label{tab:edge-tiles}%
        Arrangement of interior points for edge tiles.
        The number $h$ represents the number of interior points (coloured),
        with nearby boundary points also shown (black).
        The number $r$ represents the number of groups allocated to the tile
        during the reduction.
    }
\end{table}

\clearpage

\paragraph{Reduction step 3d: Interior points for clause tiles.}

\Cref{tab:clause-tiles} shows arrangements of interior points for each type
of clause tile.
The arrangements are such that the interior points of the clause tile can be
partitioned if and only if one of the boundary points is not included
(that is, it must be included by a neighbouring variable or edge tile,
indicating the clause will be satisfied by the corresponding literal).

\begin{lemma}
    \label{tilemma:c}
    Consider an interior and boundary point arrangement from
        \cref{tab:clause-tiles},
    or a rectilinear rotation or reflection of such an arrangement.
    Let $r \in \set{2,3}$ be the allocated number of groups,
    and let $\eps \in \PosReals$ be the scale.
    \begin{enumerate}[label=(\roman*)]
        \item
            There is no $(k,\eps)$-partition of the interior points if
            $k < r$.
        \item
            For any $(r, \eps)$-partition of the interior points,
            there is at least one boundary point that is not within
            uniform distance $\eps$ of all points in any group.
    \end{enumerate}
\end{lemma}

\begin{myproof}
    Following \cref{tilemma:v}, the conditions can be checked
    exhaustively.
\end{myproof}

\Cref{tab:clause-tiles} shows the only possible $(r,\eps)$-partitions
of the interior points, except in the third row, where a reflected
version of the first example partition is also possible.

\begin{table}[h!]
    \centering
    \includegraphics[width=\textwidth]{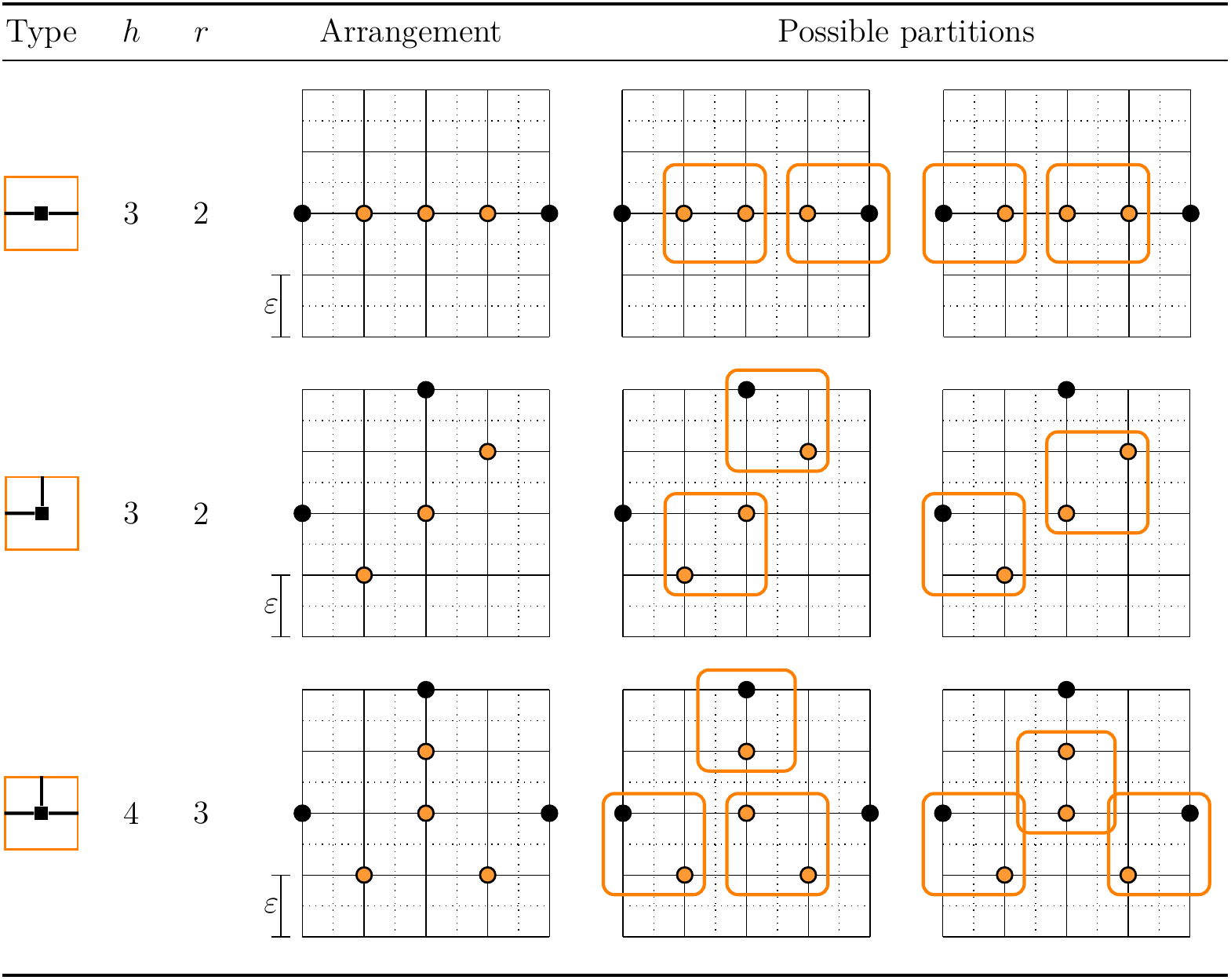}
    \caption[Arrangement of interior points for clause tiles]{%
        \label{tab:clause-tiles}%
        Arrangement of interior points for clause tiles.
        The number $h$ represents the number of interior points (coloured).
        The boundary points are also shown (black).
        The number $r$ represents the number of groups allocated to the tile
        during the reduction.
    }
\end{table}

\paragraph{Reduction step 4: Set the number of groups.}

For the number of groups, total the allocations for the interior points
of each tile
    ($r$ in \cref{tab:var-tiles,tab:clause-tiles}).
That is, set $r$ for the \UPP{} instance to 
    thrice the number of 3-occurrence variables and 3-literal clauses plus
    twice the number of edge segments, 2-occurrence variables, and 2-literal
    clauses.

\paragraph{Reduction step 5: Set the uniform diameter.}

The reduction works at any (polynomial-time computable) scale. For
concreteness, set the diameter to $1/4$, giving each tile unit width.

\paragraph{Formal summary of the reduction.}

Given an instance of \xSAT{}, that is, a restricted Boolean formula $\phi$
with variables
    $\subrange1nv$
and clauses
    $c_1 \land \cdots \land c_m$,
construct an instance of \UPP{} as described in detail in the above steps.
Namely, 
    use the points $\subrange1hx \in \Reals^2$
        as described in Reduction step~3
        (the interior points from all tiles and the boundary points
        between neighbouring tiles);
    a number of groups $r$ as described in Reduction step~4
        (the total allocated groups from all of the tiles);
    and a uniform diameter $\eps = 1/4$ as described in Reduction step~5.

\Cref{fig:full-reduction-example} shows
the full \UPP{} instances for the running examples
    (cf.~\cref{fig:xsat,fig:grid-layout,fig:boundary-points}).

\begin{table}[h!]
    \centering
    \includegraphics[width=\textwidth]{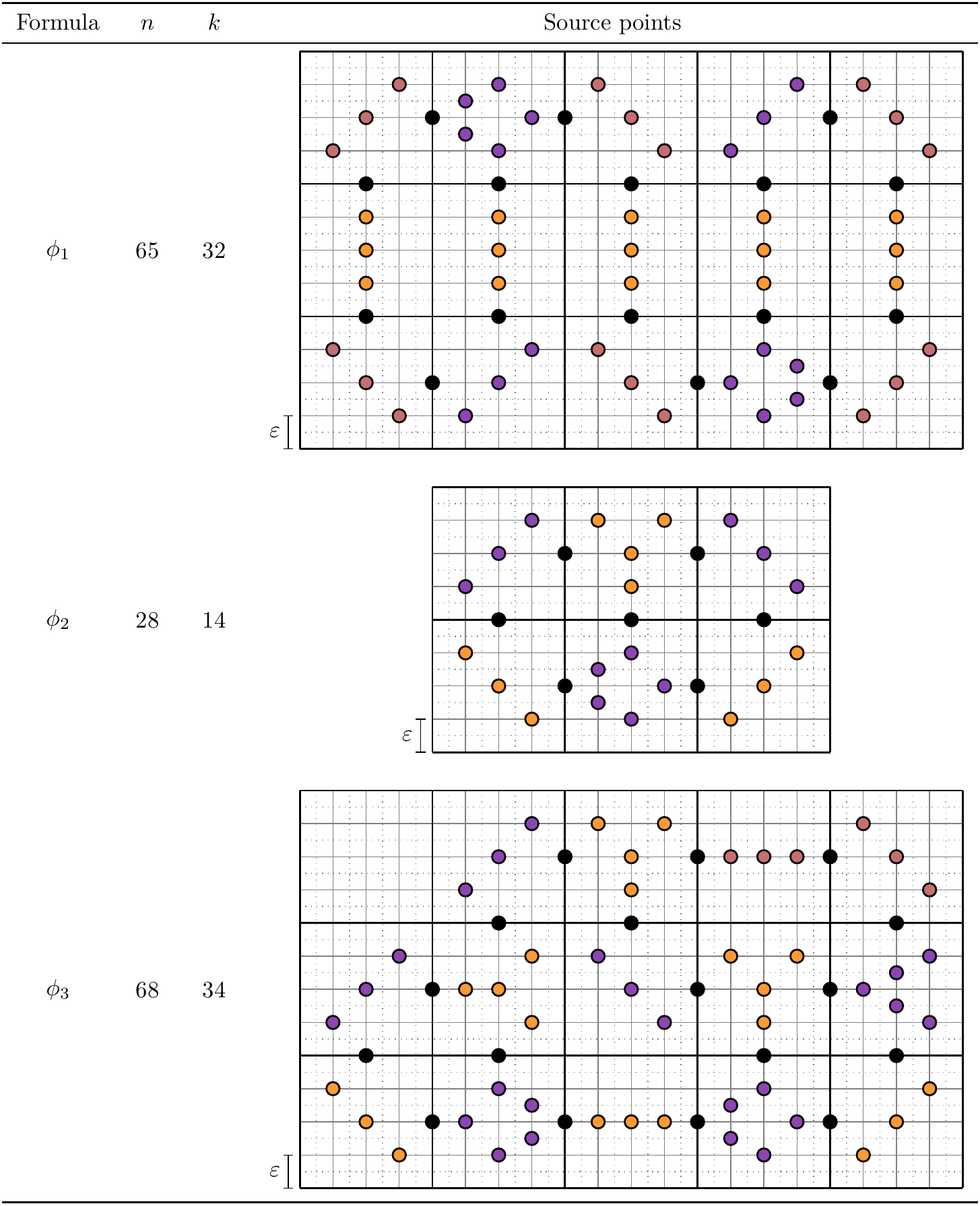}
    \caption[Full example of the reduction from \xSAT{} to \UPP{}]{%
        \label{fig:full-reduction-example}%
        Full examples of the reduction from \xSAT{} to \UPP{},
        based on \xSAT{} instances described in \cref{fig:xsat}.
        Exercise: is there an $(r,\eps)$-partition in each case?
    }
\end{table}

\clearpage
    
\paragraph{Correctness of the reduction.}

Step 1 (grid layout) runs in polynomial time \citep{Liu+1998}, and
the remaining steps run in linear or constant time.
It remains to show that the constructed instance of \UPP{} is
equivalent to the original \xSAT{} instance. That is, we must show
that $\phi$ is satisfiable if and only if there exists an
$(r, \eps)$-partition of the points $\subrange1hx$.

($\limp$):
    Suppose $\phi$ is satisfiable.
    Let $\theta$ be a satisfying truth assignment.
    Produce an $(r, \eps)$-partition of $\subrange1hx$ as follows.
    \begin{enumerate}
        \item
            Partition the interior points of each variable tile as in
            \cref{tab:var-tiles}.
            Include the positive boundary point(s) if the variable is
            assigned ``true'' in $\theta$,
            or include the negative boundary point if it is assigned
            ``false''.
        \item
            For each variable tile, follow the \emph{included} boundary
            point(s) through zero or more edge tiles to the incident clause's
            tile, partitioning the interior points of each edge tile according
            to \cref{tab:edge-tiles} such that the boundary point in the
            direction of the clause tile is included.
        \item
            Since $\theta$ is a satisfying assignment, every clause tile is
            reached in this way at least once, and thus has at least one of
            its boundary points included in the groups described so far.
            For each clause tile, partition the interior points according to
            \cref{tab:clause-tiles},
            including the remaining boundary points (if any).
        \item
            For each clause tile, follow the remaining boundary points
            through zero or more edges back to a variable tile,
            partitioning the interior points of each edge tile according to
            \cref{tab:edge-tiles} such that the boundary point in the
            direction of the variable tile is included.
    \end{enumerate}
    The final step includes exactly the boundary points of variable tiles
    that were not included in the first step.
    Thus, all interior and boundary points are included in some group.
    The number of groups is exactly in accordance with the allocated number
    of groups per tile, for a total of $r$.

($\Leftarrow$):
    Suppose there is an $(r, \eps)$-partition of the points.
    Observe the following:
    \begin{itemize}
        \item
            Since the interior points of each tile are separated from the
            tile boundaries by at least $\eps$, no group can include interior
            points from two separate tiles.
        \item
            It follows that the interior points of each tile must be
            partitioned into their allocated number of groups.
            If one tile were to use more groups, some other tile would not
            get its allocation of groups, 
            and it would be impossible to include all of its interior points 
            in the partition
            (by Lemmas~\ref{tilemma:v}(i), \ref{tilemma:e}(i), and
                \ref{tilemma:c}(i)).
        \item
            Since the boundary points have no allocated groups, each boundary
            point must be included in a group with interior points from one
            of its neighbouring tiles.
    \end{itemize}

    Now, consider each clause tile.
    By \cref{tilemma:c}(ii), there must be at least one boundary point that
    is included with the interior points of one of its \emph{neighbouring}
    tiles.
    Pick one such direction for each clause and use this to construct a
    satisfying assignment for $\phi$ as follows.
    
    In each direction, follow the sequence of zero or more edge tiles back to
    a variable tile.
    By \cref{tilemma:e}(ii), each boundary point along the sequence of edges
    must be included with the interior points of the \emph{next} edge tile in
    the sequence.
    In turn, the boundary point at the variable tile must be included with
    the interior points of the variable tile.
    If this is a positive boundary point, set this variable to ``true'' in a
    truth assignment $\theta$, and if it is a negative boundary point,
    set the variable to ``false''.
    
    This uniquely defines the truth assignment $\theta$ for all variables
    reached in this way at least once.
    If a variable is reached this way from two separate clauses, it must be
    through its two positive boundary points, since by \cref{tilemma:v}(ii),
    it is impossible for the partition to have both a negative and a positive
    boundary point included with the interior points of the variable tile.
    Since some variables may not be reached at all in this way, $\theta$ is
    not completely defined. Complete the definition of $\theta$ by assigning
    arbitrary truth values to such variables.
    
    The truth assignment $\theta$ is a satisfying assignment for $\phi$.
    Each clause is satisfied by at least one literal, corresponding to the
    variable tile that was reached through one of the clause tile's
    boundary points not grouped with the clause tile's interior points in
    the partition.

This concludes the proof of \cref{thm:xsat-reducto-upp}.
\end{myproof}

\phantomsection\label{thm:xsat-reducto-upp:end}

\clearpage
\section{Problem variations and their computational complexity}
\label{apx:variants}

In this section we discuss the computational complexity of several minor
variations of \probUPC{}.

\paragraph{Uniform vector partition is hard.}

Consider a generalisation of \probUPC{} beyond the plane, as follows.
Let $p \in \PosNats$. Given $h$ \emph{source vectors}
    $\subrange1hx \in \Reals^p$,
define an \emph{$(r,\eps)$-cover}, a list of $r$ \emph{covering vectors}
    $\subrange1ry \in \Reals^p$
such that the uniform distance between each source vector and its nearest
covering vector is at most $\eps$
    (that is, $
        \forall i \in \set{1, \ldots, h}\!,
            \exists j \in \set{1, \ldots, r}\!,
                \cdist{x_i}{y_j} \leq \eps
    $).

\begin{problem}[\UVCp]
    \label{prob:uvcp}
    Let $p \in \PosNats$.
    \emph{Uniform vector cover in $\Reals^p$} (\UVCp[p]) is a decision problem.
    Each instance comprises a collection of $h \in \Natz$ source points
        $\subrange1hx \in \Reals^p$,
    a uniform radius $\eps \in \PosReals$,
    and a number of covering points $r \in \Natz$.
    Affirmative instances are those for which there exists an
    $(r,\eps)$-cover of $\subrange1hx$.
\end{problem}

\begin{theorem}
    If $p \geq 2$, then \UVCp[\,p] is \NP-complete.
\end{theorem}

\begin{myproof}
    ($\UPC \reducto \UVCp$):
    Let $p \geq 2$.
    Embed the source points from the \UPC{} instance into the first two
    dimensions of $\Reals^p$ (leaving the remaining components zero).
    If there is an $(r,\eps)$-cover of the 2-dimensional points, embed it
    similarly to derive a $p$-dimensional $(r,\eps)$-cover.
    Conversely, if there is a $p$-dimensional $(r,\eps)$-cover, truncate it
    to the first two dimensions to derive an $(r,\eps)$-cover.
    
($\UVCp \in \NP$):
    Use an $(r,2\eps)$-partition (suitably generalised to $p$ dimensions)
    as a polynomial-time verifiable certificate.
    Such a certificate is appropriate along the lines of the proof of 
        \cref{thm:three-perspectives}.
    (An arbitrary $(r,\eps)$-cover is unsuitable as a certificate along the
    lines of \cref{fn:certification}.)
\end{myproof}

\begin{remark}
    By a $p$-dimensional generalisation of \cref{thm:three-perspectives},
    $p$-dimensional generalisations of \probUPP{} and
    \pref{prob:usgcp}{\usgCP} are also \NP-complete for $p \geq 2$.
\end{remark}

\paragraph{Uniform scalar partition is easy.}

On the other hand, \UVCp[1], which could be called \emph{uniform scalar
cover}, is in \Poly{}.
A minimal cover can be constructed using a greedy algorithm with runtime
    $\BigO(h\log h)$.
An $(r,\eps)$-cover exists if and only if there are at most $r$ scalars
in the result.

\begin{algorithm}[Optimal uniform scalar cover]\label{algo:usc}
    Proceed:
    \begin{algorithmic}[1]
    \Procedure{UniformScalarCover}{%
            $h\in\Natz$,
            $\eps\in\PosReals$,
            $\subrange1hx\in\Reals$%
        }
        \State $\subrange1h{x'} \gets \subrange1hx$,
                                        sorted in non-decreasing order.
        \State $j \gets 0$
        \For{$i = \range1h$}
            \If{$j = 0$ or $x'_i > y_j + \eps$}
                \State $j \gets j+1$
                \State $y_j \gets x'_i + \eps$
            \EndIf
        \EndFor
        \State \Return $\subrange1jy$
    \EndProcedure
    \end{algorithmic}
\end{algorithm}

\begin{theorem}[\cref{algo:usc} correctness]
    Let $h\in\Natz$, $\eps\in\PosReals$, and $\subrange1hx\in\Reals$.
    Let $\subrange1ry = \CALL{UniformScalarCover}(h,\eps,\subrange1hx)$.
    Then (i)~$\subrange1ry$ is an $(r,\eps)$-cover of $\subrange1hx$;
    and (ii)~no $(k,\eps)$-cover exists for $k < r$.
\end{theorem}

\begin{myproof}
(i):
    After each iteration $i = \range1h$, if the if branch was entered,
    then
        $
            \cdist{x'_i}{y_j}
            = \cdist{x'_i}{(x'_i + \eps)}
            = \eps
        $.
    If not, that is, if $x'_i \leq y_j+\eps$, let $k$ be the last iteration
    in which $j$ was increased. Then $x'_i \geq x'_k = y_j - \eps$.
    In summary, $y_j - \eps \leq x'_i \leq y_j+\eps$, as required.

(ii): For $j = \range1r$, let $\xi_j = x'_i$ from the iteration in which
    $y_j$ was defined.
    Then for $j \neq k$, $\cdist{\xi_j}{\xi_k} \geq 2\eps$.
    No covering scalar could be within distance $\eps$ of any two of these
    source scalars. Thus at least $r$ covering scalars are required to cover
    $\subrange1r\xi$, and, in turn, $\subrange1hx$.
\end{myproof}

\paragraph{Clique partition on `square penny' graphs is hard.}

\citet{Cerioli+2004,Cerioli+2011} showed that clique partition is
\NP-complete in a restricted variant of of unit disk graphs called
\emph{penny graphs}.
A penny graph is a unit disk graph in which the Euclidean distance between
source points is \emph{at least} $\eps$, evoking the contact relationships
among non-overlapping circular coins.

Our reduction (\cref{apx:complexity}) happens to produce a set of source
points for a unit square graph satisfying a uniform distance version of
this condition.
Therefore, our proof shows that clique partition remains \NP-complete in this
special family of graphs as well.

\clearpage
\section{Bounding proximate rank with biasless units is still hard}
\label{apx:unbiased}

In this appendix, we consider a simpler neural network architecture with
unbiased hidden units (such that unit clustering takes place in one
dimension).
In the main paper we proved the hardness of \probPAR{} using a reduction from
\probUPC{}, a planar clustering problem.
While a one-dimensional variant of \probUPC{} is in $\Poly$
    (\cref{apx:variants}),
we show that in a \emph{biasless} architecture, bounding proximate rank
is still \NP-complete, by reduction from a subset sum problem.

This biasless architecture is of limited practical interest---rather, it
provides a minimal illustration of a ``second layer of complexity'' in
computing the proximate rank.
Roughly: computing an optimal approximate partition of the units alone is not
sufficient---one must seek an approximate partition that \emph{jointly}
maximises the numbers of units that can be eliminated (1)~by approximate
merging \emph{and} (2)~with total outgoing weight near zero
    (cf.~\cref{algo:greedy-prank-bound}).
The reduction of subset sum to bounding proximate rank in biasless networks
is chiefly based on objective~(2).
The reduction is not specific to biasless networks---a similar reduction is
possible for biased architectures.\footnote{%
    The same holds for architectures with multi-dimensional inputs and outputs
    (cf.~\cref{apx:multidim}).
    The reduction could also be achieved using an architecture with
    biased hidden units that have shared incoming weights.
}

\paragraph{Biasless hyperbolic tangent networks.}

Formally, consider an architecture with one input unit, one biased output
unit, and one hidden layer of $h\in\Natz$ \emph{unbiased} hidden units with
the hyperbolic tangent nonlinearity.
The weights of the network are encoded in a parameter vector in the format
    $u = (a_1,b_1,\ldots,a_h,b_h) \in \U_h = \Reals^{2h}$,
where for each hidden unit $i=\range1h$ there is an outgoing weight
$a_i\in\Reals$ and an incoming weight $b_i\in\Reals$.
Thus each parameter $u \in \U_h$ indexes a mathematical function
    $f_u : \Reals \to \Reals$
such that
    $f_u(x) = \sum_{i=1}^h a_i \tanh(b_i x)$.

The notions of functional equivalence, lossless compressibility, rank, and
proximate rank from the main paper generalise straightforwardly
to this setting. The analogous reducibility conditions are:
\begin{enumerate}[label=(\roman*)]
    \item $a_i=0$ for some~$i$, or
    \item $b_i=0$ for some~$i$, or
    \item $b_i=b_j$ for some $i \neq j$, or
    \item $b_i=-b_j$ for some $i \neq j$.
\end{enumerate}

\paragraph{Proper subset sum zero.}

The subset sum problem concerns finding a subset of a collection of positive
integers with a given sum
    \citep{Karp1972,Garey+Johnson1979}.\footnote{%
        \citet{Karp1972} studied the subset sum problem under the name
        ``knapsack'' and used a slightly different formulation allowing
        non-positive integers in the collection and target.
    }

\begin{problem}[\SSum]
    \label{prob:ssum}
    \emph{Subset sum} (\SSum) is a decision problem.
    Each instance comprises
        a collection of $n \in \PosNats$ positive integers,
            $\subrange1nx \in \PosNats$
        and a target integer $T \in \PosNats$.
    Affirmative instances are those for which there exists a subset
        $S \subseteq \setrange1n$
    such that
        $\sum_{i\in S} x_i = T$.
\end{problem}

\begin{theorem}[\citealp{Karp1972,Garey+Johnson1979}]
    \probSS{} is \NP-complete.
\end{theorem}

We pursue a reduction to \probUPAR{} from the following \NP-complete variant:

\begin{problem}[\SSZ]
    \label{prob:ssumzero}
    \emph{Proper subset sum zero} (\SSZ) is a decision problem.
    Each instance comprises a collection of $n \in \PosNats$ positive or
    negative integers,
        $\subrange1nx \in \mathbb{Z} \setminus \set{0}$.
    Affirmative instances are those for which there exists a non-empty,
    proper subset
        $S \subsetneq \setrange1n$
    such that
        $\sum_{i\in S} x_i = 0$.
\end{problem}

\begin{theorem}
    \label{thm:ssumzero-npc}
    \probSSZ{} is \NP-complete.
\end{theorem}

\begin{myproof}
($\SSZ\in\NP$):
    The subset $S$ serves as its own certificate.
    Its validity can be verified in polynomial time by checking that
        $\set{} \subsetneq S \subsetneq \setrange1n$
    and $\sum_{i \in S} x_i = 0$.
    
($\SSum \reducto \SSZ$):
    Consider an \SSum{} instance with collection $\subrange1nx \PosNats$
    and target $T \in \PosNats$.
    First, rule out the case $\sum{i=1}^n x_i = T$
        (if so, return a trivial affirmative instance of \SSZ{}).
    Otherwise, let $x_{n+1} = -T$ and return the \SSZ{} instance
    with collection
        $\subrange1{n+1}x \in \mathbb{Z} \setminus \set{0}$.
    Now, consider an affirmative instance of $\SSum$ with subset
        $S \subseteq \setrange1n$.
    If $S = \setrange1n$ then the new instance is affirmative by
    construction.
    Otherwise, $S' = S \cup \set{n+1}$ is a non-empty, proper subset of
    $\setrange1{n+1}$ such that
        $\sum_{i \in S'} x_i = \left(\sum_{i\in S} x_i\right) - T = 0$.
    Conversely, if the constructed instance is affirmative (aside from in
    the trivial case) with subset $S'$, then
        $n+1 \in S'$ since $\subrange1nx > 0$.
    It follows that $S = S'\setminus\set{n+1}$ satisfies
        $
            \sum_{i \in S} x_i
            = \left(\sum_{i \in S'} x_i \right) - x_{n+1}
            = 0 - (-T)
            = T
        $.
\end{myproof}

\paragraph{Complexity of unbiased proximate rank.}

We formalise the problem of bounding the proximate rank in the biasless
architecture as follows.

\begin{problem}[\uPAR]
    \label{prob:upar}
    \emph{Bounding biasless proximate rank} (\uPAR) is a decision
    problem.
    Each instance comprises
        a number of hidden units $h \in \Natz$,
        a parameter $u\in\U_h$,
        a uniform radius $\eps\in\PosReals$,
    and a maximum rank $r\in\Natz$.
    Affirmative instances are those for which
        $\prank\eps(u) \leq r$.
\end{problem}

\begin{theorem}
    \label{thm:upar-npc}
    \Pref{prob:upar}{\uPAR} is \NP-complete.
\end{theorem}

\begin{myproof}
    Since \SSZ{} is \NP-complete (\cref{thm:ssumzero-npc}), it suffices to
    show $\SSZ \reducto \uPAR$ and $\uPAR \in \NP$.

($\SSZ \reducto \uPAR$, reduction):
    Consider an instance of \probSSZ{} with
        $\subrange1nx \in \mathbb{Z} \setminus \set{0}$.
    Construct an instance of \probUPAR{} in $\BigO(n^2)$ time as follows
    (see \cref{fig:prank-biasless-reduction-example} for an example).
    Set $h = n^2$, conceptually dividing the $n^2$ hidden units into
    $n$~groups of $n$~units.
    Construct a parameter
        \begin{align*}
            u &= (
                a_{1,1}, b_{1,1},
                \ldots,
                a_{1,n}, b_{1,n},
                a_{2,1}, b_{2,1},
                \ldots,
                a_{n,n}, b_{n,n}
            ) \in \U_{n^2},
        \intertext{%
            where for each unit $j=\range1n$ of each unit group
            $i=\range1n$:\footnotemark{}
        }
            a_{i,j} &=  n \cdot x_{\modulo(i + j - 1, n)},
        \\  b_{i,j} &=  (2i-1) + \frac{j-1}{n-1}.
        \end{align*}
    That is, the outgoing weights for units in group $i$ are just the
    collection $\subrange1nx$ cycled to start from $x_i$, and scaled by a
    factor of $n$;
    and the incoming weights for units in group $i$ are evenly spaced values
    in the interval from $2i-1$ to $2i$.

    \footnotetext{%
        Here, $\modulo(i, n)$ represents $i$ shifted by some multiple of $n$
        into the range $\setrange1n$.
        For example,
            $\modulo(-2,3)=\modulo(1,3)=\modulo(4,3)=1$
        and 
            $\modulo(-3,3)=\modulo(0,3)=\modulo(3,3)=\modulo(6,3)=3$.
    }

    Moreover, set
        $\eps = \frac12\frac{n-2}{n-1}$,
    such that an interval of width $2\eps$ can cover $n-1$ of the incoming
    weights from any group, but not all $n$, and no two units from different
    groups.
    Finally, set
        $r = 2n-1$,
    that is, one less than enough to allocate two compressed units to each of
    the $n$ groups.

\begin{figure}[!ht]
    \centering
    \includegraphics[]{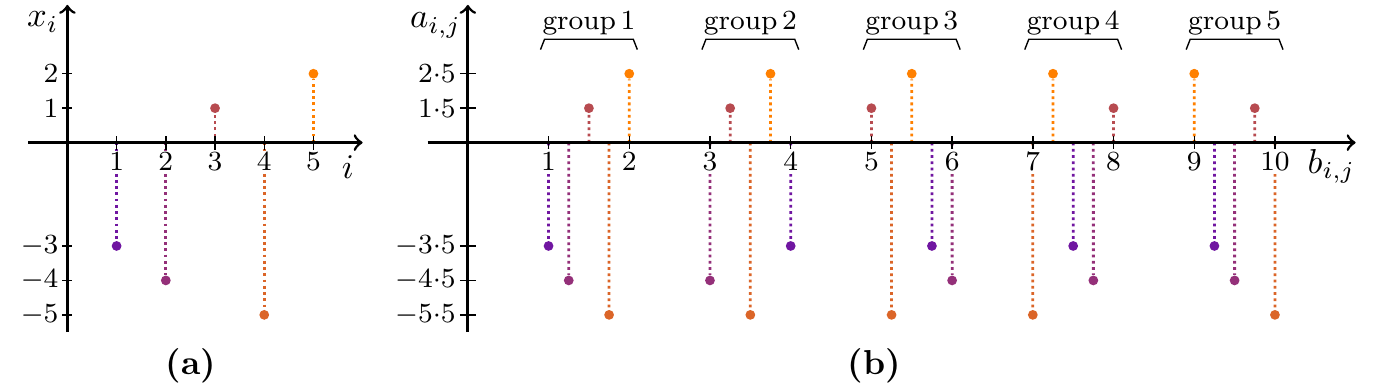}
    \caption{\label{fig:prank-biasless-reduction-example}%
        Illustrative example of the parameter construction.
        \textbf{(a)}~A \probSSZ{} instance with collection
            $\subrange15x = -3, -4, 1, -5, 2$.
        \textbf{(b)}~Scatter plot of the incoming and outgoing weights of
        the $5$ groups of $5$ units of the constructed parameter
            $u \in \U_{25}$.
        Moreover, $\eps = 3/8$ and $r = 9$.
    }
\end{figure}

($\SSZ \reducto \uPAR$, equivalence):
    It remains to show that the constructed instance of \uPAR{} is
    affirmative if and only if the given instance of \SSZ{} is affirmative,
    that is, there exists a nonempty proper subset with zero sum
    if and only if the constructed parameter has $\prank\eps(u) \leq r$.
    They key idea is that a nearby sufficiently-compressible parameter
    must have a set of units that are merged and then eliminated with
    total outgoing weight zero, corresponding to a zero-sum subset
    (cf.~\cref{fig:prank-biasless-reduction-equivalence}).

($\limp$):
    Let $S \subsetneq \setrange1n$ be a nonempty proper subset such that
        $\sum_{k \in S} x_k = 0$.
    Pick~$i$ such that $i \in S$ and $\modulo(i+(n-1),n) \notin S$.
    Let $S_i = \Set{j \in \setrange1n}{\modulo(i+j-1, n) \in S}$.
    That is, if unit $j$ of group $i$ was assigned outgoing weight based on
    integer $x_k$ during the reduction, then $j \in S_i \liff k \in S$.
    Then $n \notin S_i$ since $\modulo(i+n-1,n) \notin S$ by assumption.
    Thus, one can approximately merge the group~$i$ units in $S_i$ into a
    single unit with incoming weight
        $2i:set non - 1 + \eps$
    and outgoing weight
        $\sum_{k \in S} n \cdot x_k = n \cdot 0 = 0$.
    Moreover, $1 \in S_i$ since $\modulo(i+1-1,n) = i \in S$, so 
    the remaining units $j \notin S_i$ in the same group are approximately
    mergeable into a unit with incoming weight $2i - \eps$.
    Similarly, the units of each remaining group are approximately mergeable
    into two units per group.
    This results in a total of $2n$ approximately merged units, at least one
    of which has outgoing weight zero.
    It follows that
        $\prank\eps(u) \leq 2n-1 = r$
    (cf.~proof of \cref{thm:algo:greedy-prank-bound}).

\clearpage

($\Leftarrow$):
    Suppose $\prank\eps(u) \leq r$, with $u^\star \in \cnbhd\eps{u}$ such
    that $\rank(w^\star) = r^\star \leq r = 2n-1$.
    In general, the only ways that $u^\star$ could have reduced rank compared
    to $u$ are the following:
    \begin{enumerate}
        \item
            Some incoming weight $b_{i,j}$ could be perturbed to zero,
            allowing its unit to be eliminated.
        \item
            Two or more units from one unit group with (absolute)
            incoming weights within $2\eps$ could have these weights
            perturbed to be equal (in magnitude), allowing them to be
            merged.
        \item
            Two or more units from two or more groups could be similarly
            merged.
        \item
            Some group of $m \geq 1$ units, merged through the above options,
            with total outgoing weight within $m\eps$ of zero, could have
            their outgoing weights perturbed to make the total zero.
    \end{enumerate}
    By construction, all $b_{i,j} \geq 1 > \eps$, so we can rule out
    option~(1), and we can rule out any merging involving negative incoming
    weights for options~(2) or~(3).
    Then, since the incoming weights of unit groups are separated by at
    least $1 > 2\eps$, option~(3) is ruled out.
    This leaves (the positive version of) option~(2) and option~(4)
    responsible for the reduction in rank.

    Since there are $n$ unit groups but the proximate rank is at most
    $2n-1$, there must be some group where after merging and elimination,
    there is at most \emph{one} unit left. Let $i$ designate this unit group.
    Units $1$ and $n$ of unit group $i$ have incoming weights separated by
        $1 > 2\eps$,
    so it is not possible to merge all of the units in this group into a
    single new unit.
    They must be merged into two or more new units, all but (at most) one of
    which are then eliminated using option~(4) above.
    Let $S_i$ designate one such set of units within group $i$ to be merged
    and eliminated.
    It follows that 
    \begin{equation*}
        \abs\left(\sum_{j \in S_i} a_{i,j}\right)
        \leq
        \eps \cdot \cardinality{S_i}
        < \eps \cdot n
        < n
        .
    \end{equation*}
    Now, let $S = \Set{\modulo(i+j-1,n)}{j \in S_i}$. That is, if unit $j$ of
    group $i$ was assigned outgoing weight based on integer $x_k$ during the
    reduction, then $k \in S \liff j \in S_i$.
    Then the set $S$ is a nonempty proper subset of $\setrange1n$ by
    construction.
    Moreover, the corresponding integers sum to an integer with magnitude
    less than 1, that is, to zero:
    \begin{equation*}
        \abs\left( \sum_{k \in S} x_k \right)
        = \frac1n \abs\left(
              \sum_{j \in S_i} n \cdot x_{\modulo(i+j-1,n)}
          \right)
        = \frac1n \abs\left( \sum_{j \in S_i} a_{i,j} \right)
        < 1
        .
    \end{equation*}
    
\begin{figure}[!ht]
    \centering
    \includegraphics[]{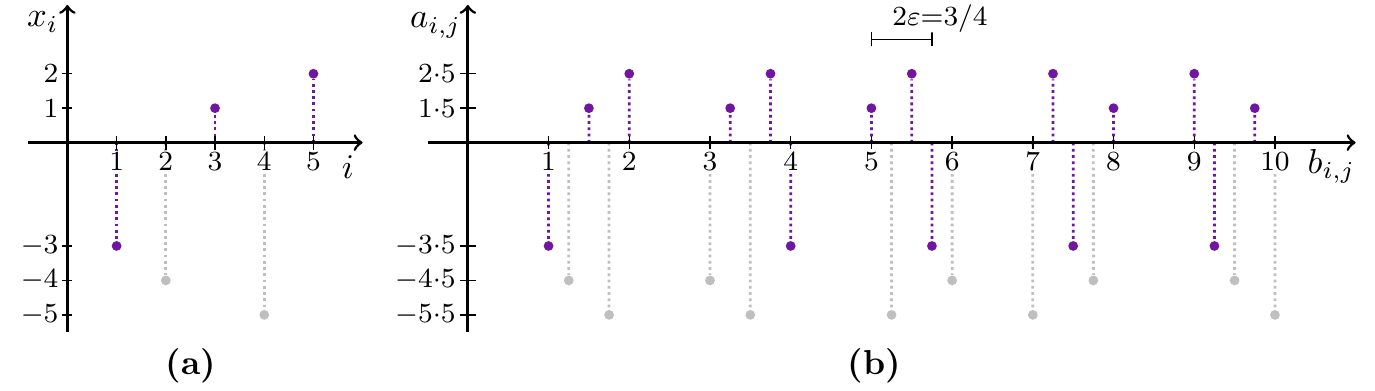}
    \caption{\label{fig:prank-biasless-reduction-equivalence}%
        Illustrative example of the correspondence between zero-sum subsets
        and eliminability of approximately-merged units.
        \textbf{(a)}~An affirmative \probSSZ{} instance 
            (cf.~\cref{fig:prank-biasless-reduction-example})
        with collection
            $\subrange15x = -3, -4, 1, -5, 2$
        and zero-sum subset $S = \set{1,3,5}$
            (observe $x_1+x_3+x_5=0$).
        \textbf{(b)}~In unit group~3 of the constructed parameter,
            units
                $1$, $3$, and $4$
            have outgoing weights
                $5 \cdot x_3$, $5 \cdot x_5$, and $5 \cdot x_1$,
            respectively,
            and incoming weights within $2\eps = \frac34$.
            These units can be approximately merged and then the merged unit
            can be eliminated, since it has outgoing weight
                $5\cdot(x_1+x_3+x_5)=0$.
    }
\end{figure}

($\uPAR \in \NP$):
    Consider an instance
        $h, r \in \Natz$,
        $\eps \in \PosReals$,
    and $u = (a_1, b_1, \ldots, a_h, b_h) \in \U_h$.
    Use as a certificate a partition $\subrange1J\Pi$
    of $\Set{i\in\setrange1h}{\abs(b_i) > \eps}$,
    such that (1)~for each $\Pi_j$, for each $i, k \in \Pi_j$,
        $\abs(\abs(b_i) - \abs(b_k)) \leq2\eps$;
    and (2)~at most $r$ of the $\Pi_j$ satisfy
        $\sum_{i\in\Pi_j} \sign(b_i) \cdot a_i > \eps \cdot \cardinality{\Pi_j}$.
    The validity of such a certificate can be verified in polynomial time by
    checking each of these conditions directly.
    Such a certificate exists if and only if $\prank\eps(u) \leq r$, by similar
    reasoning as used in the proof of \cref{thm:par-npc} ($\PAR\in\NP$).
\end{myproof}

\clearpage
\section{Hyperbolic tangent networks with multi-dimensional inputs and outputs}
\label{apx:multidim}

In the main paper we consider single-hidden-layer hyperbolic tangent networks
with a single input unit and a single output unit.
Our algorithms and results generalise to an architecture with multiple input
units and multiple output units, with some minor changes.

Consider a family of fully-connected, feed-forward neural network
architectures with $n\in\PosNats$ input units, $m \in \PosNats$ biased linear
output units, and a single hidden layer of $h\in\Natz$ biased hidden units
with the hyperbolic tangent nonlinearity.
The weights and biases of the network are encoded in a parameter vector in
the format
    $w = (\abcrange, d) \in \W[n,m]_h = \Reals^{(n+m+1)h+m}$,
where for each hidden unit $i = \range1h$ there is
    an \emph{outgoing weight vector} $a_i \in \Reals^m$,
    an \emph{incoming weight vector} $b_i \in \Reals^n$,
    and a bias $c_i \in \Reals$;
and $d \in \Reals^m$ is a vector of output unit biases.
Thus each parameter $w \in \W[n,m]_h$ indexes a function
    $f_w : \Reals^n \to \Reals^m$ such that
    $f_w(x) = d + \sum_{i=1}^h a_i \tanh(b_i \cdot x + c_i)$.

The above notation is deliberately chosen to parallel the case $n=m=1$
considered in the main paper. This makes the generalisation of our results to
the case $n,m\geq1$ straightforward. First, replace all mentions of the
scalar incoming and outgoing weights with incoming and outgoing weight
\emph{vectors}.
It remains to note the following additional changes.

\begin{enumerate}
    \item
        The algorithms and proofs commonly refer to the sign of an incoming
        weight. For $b \in \Reals^n$ define $\sign(b) \in \set{-1, 0, +1}$
        as the sign of the first nonzero component of $b$, or zero if $b=0$.
        Use this generalised sign function throughout when $n > 1$.
    \item
        Sorting the (signed) incoming weight and bias pairs of the
        hidden units is a key part of
            \cref{algo:lossless-compression,algo:rank}.
        For pairs of the form $(b, c) \in \Reals^n \times \Reals$,
        lexicographically sort by the tuple $(\subrange1n{b}, c)$.
    \item
        The reducibility conditions were proven by \citet{Sussmann1992}
        in the case $n \geq 1$ and $m=1$.
        The conditions also hold for $m \geq 1$
        \citetext{%
            see
            \citealp{Fukumizu1996};
            or
            \citealp[\S{}A]{Farrugia2023}%
        }.
    \item
        The construction of the centre of a bounding rectangle in the
        proof of \cref{thm:par-npc} straightforwardly generalises to
        the construction of the centre of a bounding right cuboid.
\end{enumerate}

For \cref{thm:par-npc}, the proof that $\UPC \reducto \PAR$ requires no
generalisation, because it suffices to construct instances of \probPAR{}
in the special case $n=m=1$ to prove that this more general problem is
hard.
To see directly that \probPAR{} remains hard in networks with $n$ input
units, consider that additional weights can be assigned zero in an extended
construction---compare with \UVCp[n+1] (\pref{prob:uvcp}{\UVCp} discussed in
\cref{apx:variants}).\footnote{%
    Each hidden unit translates to an ($n{+}1$)-dimensional vector, with
    $n$ incoming weights and $1$~bias weight.
    See also \cref{apx:unbiased} for a discussion of the biasless case,
    where bounding proximate rank is still hard
        (\cref{thm:upar-npc})
    even though $\UVCp[1]$ is easy
        (\cref{apx:variants}).
}

\end{document}